\pdfoutput=1
\RequirePackage{fix-cm}
\documentclass[10pt,twocolumn,letterpaper]{article}
\usepackage[margin=0.8in,columnsep=0.3in]{geometry}
\usepackage[T1]{fontenc}
\usepackage{times}
\usepackage{amsmath,amssymb,amsthm}
\usepackage{graphicx,booktabs,multirow,pifont}
\usepackage[table]{xcolor}
\usepackage{arydshln}
\usepackage{pdflscape}
\usepackage[font=small,labelfont=bf]{caption}
\usepackage[numbers,sort&compress]{natbib}
\usepackage{microtype}
\definecolor{linkblue}{RGB}{36,84,119}
\usepackage[colorlinks,allcolors=linkblue,bookmarksnumbered]{hyperref}

\newtheorem{proposition}{Proposition}
\newcommand{\PaperTitle}{TALD: Transferability-Aware Logit Disentanglement for Cross-Modal Knowledge Distillation}
\newcommand{\affmark}[1]{\texorpdfstring{\textsuperscript{#1}}{}}
\newcommand{\AuthorNames}{Hui Li\affmark{1}, Pengfei Yang\affmark{1}, Xinyi Feng\affmark{1}, Junhong Liu\affmark{2}, Tianyang Zheng\affmark{1}, Hui Zeng\affmark{1}, Le Dong\affmark{1}, Kaibing Zhang\affmark{3}, Mengting Zhao\affmark{1}, Quan Wang\affmark{1}}
\newcommand{\Affiliations}{\affmark{1}Xidian University\qquad \affmark{2}Beihang University\qquad \affmark{3}Xi'an Polytechnic University}
\newcommand{\ContactEmail}{\nolinkurl{gray1y@stu.xidian.edu.cn}}

\title{\bfseries\PaperTitle}
\author{\parbox{0.96\textwidth}{\centering
  \AuthorNames\\[3pt]
  {\small\Affiliations}\\
  {\small\ContactEmail}}}
\date{}
\hypersetup{pdftitle={\PaperTitle},pdfauthor={\AuthorNames},pdflang={en}}

\begin{document}
\maketitle
\begin{abstract}
	Cross-modal knowledge distillation can incur negative transfer when source-modality supervision conflicts with the target modality's discriminative characteristics. We propose TALD, a two-stage framework for Transferability-Aware Logit Disentanglement in class-aligned logit space. Guided by task-dependent self-benefiting and cross-modal exclusivity criteria, a Residual Disentangler decomposes logits from a teacher and an independently trained target-modality reference student into modality-general and modality-specific components. A Sample-Adaptive Modulator regulates decomposition constraints and sample-specific fusion weights. The learned modality-general components are then adaptively fused to supervise the final student. Across five benchmarks spanning vision, audio, and text, TALD achieves the largest mean accuracy gain and highest positive-transfer frequency among the compared methods. Across ten transfer directions with five runs each, TALD improves over training without distillation in 84\% of evaluations, compared with at most 70\% for the baselines, while offering a favorable accuracy--efficiency trade-off. We plan to release the code.
\end{abstract}

\section{Introduction}

\begin{figure}[t]
	\centering
	\includegraphics[width=\linewidth]{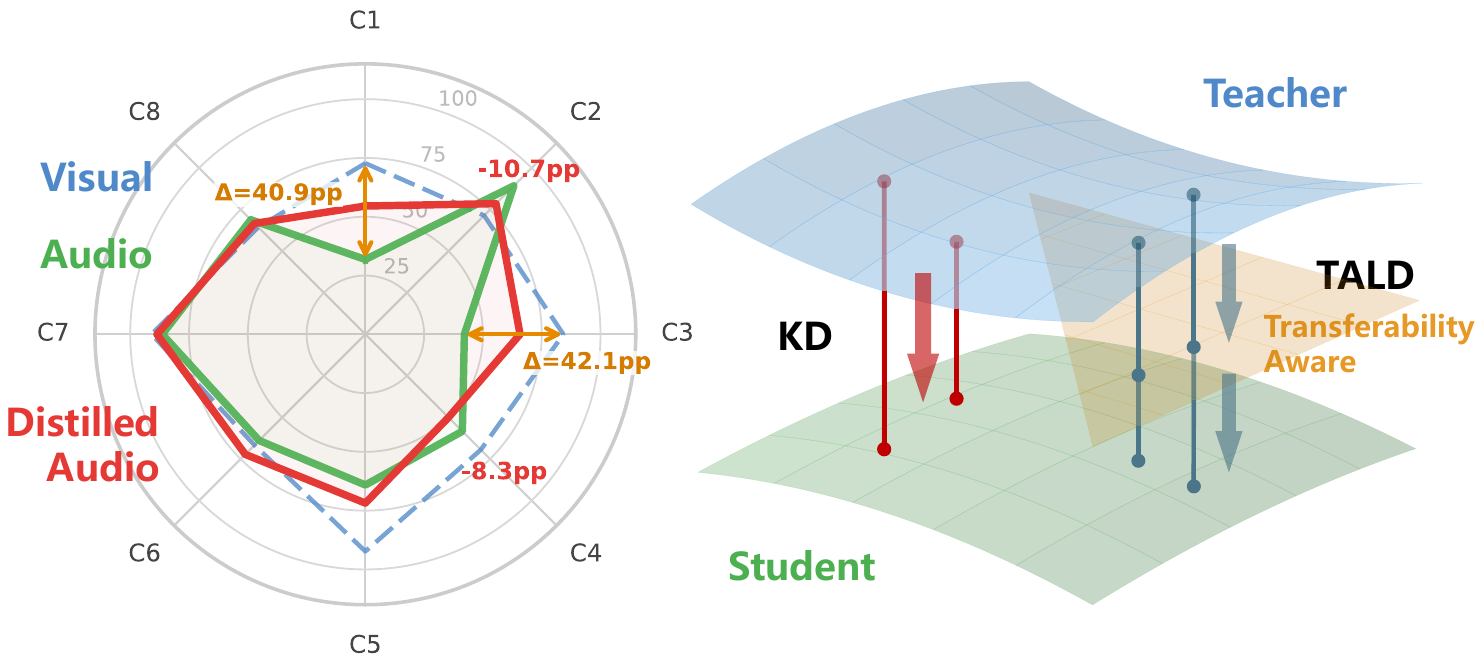}
	\caption{Class-structured negative transfer and the transferability-aware view (RAVDESS, V$\to$A; single run). The visual and audio baselines differ in per-class accuracy by up to about 42 percentage points (pp), and the KD student falls below the audio baseline on several classes (e.g., C2, C4). Conventional KD uses the full teacher response, while TALD constructs its distillation target by adaptively fusing learned modality-general components from the teacher and reference student.}
	\label{fig:motivation}
\end{figure}

Cross-modal knowledge distillation (CMKD) transfers knowledge from a source-modality teacher to a target-modality student~\cite{gupta2016cross, garcia2018modality}. Although the modalities share class labels, their discriminative cues can differ, and knowledge useful to the teacher may not benefit the student~\cite{mfh, c2kd}. As illustrated in Figure~\ref{fig:motivation}, the modalities exhibit different class-wise strengths, and conventional KD can underperform the target-modality baseline on some classes. This motivates distillation targets that retain useful cross-modal information while accounting for the target modality's own discriminative characteristics.

The Modality Focusing Hypothesis (MFH)~\cite{mfh} links distillation effectiveness to the proportion of modality-general discriminative information retained by the teacher. Existing methods address modality discrepancies through teacher adaptation~\cite{mst-distill, c2kd} or feature filtering and reconstruction~\cite{mfh, mst-distill}. In particular, MGDFR~\cite{mfh} uses channel saliency as a proxy for transferability to reconstruct high-dimensional teacher representations. Such feature-space processing provides an indirect route to shaping class-level supervision and can incur substantial computational overhead. This motivates a more compact space in which the classification effects of the transferred components can be modeled directly.

Logits provide such a space. For a shared classification task, corresponding dimensions represent the same classes across modalities~\cite{sun2024logit}, allowing the classification effects of logit components to be evaluated directly through the task loss. We therefore formulate disentanglement in this class-aligned space as a residual decomposition, subtracting a learned modality-specific component from the original logits to obtain the modality-general component. This parameterization ties the two components to the original class scores through exact additive reconstruction. Further analyses and visualizations of logit-space structure and the residual formulation are provided in Appendix~\ref{appendix:moti}.

We propose Transferability-Aware Logit Disentanglement (TALD), a two-stage CMKD framework that constructs distillation targets from a frozen teacher and an independently trained target-modality reference student. In Stage~1, a Residual Disentangler learns a criterion-guided logit decomposition, while a Sample-Adaptive Modulator regulates the constraints. In Stage~2, the learned modality-general components are fused using the Modulator's sample-specific weights to supervise the final student. Our main contributions are:

\begin{itemize}
	\item We formulate task-dependent self-benefiting and cross-modal exclusivity criteria to characterize the desired classification effects of logit components within and across modalities, providing behavioral guidance for disentanglement.

	\item We propose TALD, a two-stage CMKD framework that integrates residual logit decomposition with sample-adaptive constraint regulation and fusion of the teacher's and reference student's learned modality-general components.

	\item Extensive experiments on five cross-modal benchmarks demonstrate TALD's effectiveness in mitigating negative transfer while achieving competitive accuracy and a favorable accuracy--efficiency trade-off.
\end{itemize}

\section{Related Work}

\subsection{Multimodal Representation Disentanglement}

Disentangled representation learning aims to separate distinct factors of variation~\cite{drl}. Shared-private formulations have been studied in domain adaptation~\cite{Bousmalis} and image translation~\cite{Lee_2018, Huang_2018}. In multimodal representation learning, MISA~\cite{misa} learns modality-invariant and modality-specific representations for sentiment analysis, while DMVAE~\cite{dmvae} separates shared and private latent variables in a generative model. Other methods use adversarial objectives to separate shared and private features~\cite{yang2022disentangled} or perform graph distillation within decoupled feature spaces~\cite{Li_2023_CVPR}. Complementary theoretical work examines the conditions and trade-offs involved in separating shared and modality-specific information~\cite{disentangledssl}, as well as the role of cross-modal distillation in reducing feature interference during multimodal fusion~\cite{chaudhuricloser}. A shared-private partition does not by itself establish the benefit of each component for a target-modality student. TALD uses task-dependent criteria based on the intended within- and cross-modal classification effects of logit components to guide decomposition and target construction.

\subsection{Cross-Modal Knowledge Distillation}

Knowledge distillation transfers knowledge by matching teacher outputs~\cite{kd} or intermediate features~\cite{fitnets}, and has been extended to cross-modal settings to improve a target-modality student using knowledge from other modalities~\cite{gupta2016cross, garcia2018modality, thoker2019cross, hu2020knowledge}. Related work also explores response decoupling for sentiment analysis with missing modalities, using a teacher with complete multimodal inputs to guide a student with incomplete inputs~\cite{corrkd}. The Modality Focusing Hypothesis (MFH)~\cite{mfh} links distillation effectiveness to the proportion of modality-general discriminative information retained by the teacher, motivating saliency-based feature filtering. MST-Distill~\cite{mst-distill} combines multimodal and auxiliary unimodal teachers through feature masking and instance-level routing. C$^2$KD~\cite{c2kd} selects samples based on consistency in logit rankings and uses bidirectional adaptation to reduce the modality gap. FD-CMKD~\cite{fd-cmkd} applies different alignment losses to low- and high-frequency feature components. TALD uses a unimodal source teacher and a target-modality reference student, constructing its target through criterion-guided residual decomposition and adaptive fusion in class-aligned logit space.

\begin{figure*}[t]
	\centering
	\includegraphics[width=0.9\linewidth]{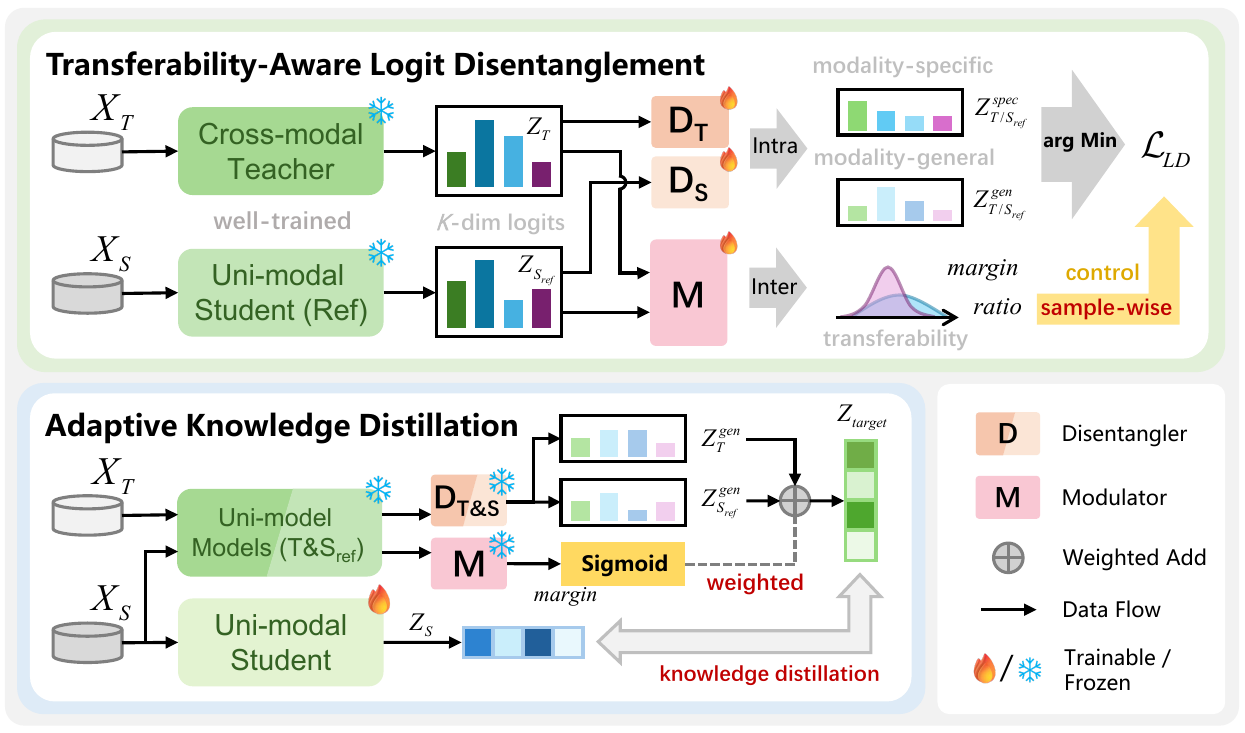}
	\caption{Overview of the TALD framework. In the first stage,
		the Disentangler decomposes teacher and reference student logits
		into modality-general and modality-specific components, while
		the Modulator predicts sample-specific margins and
		contribution-ratio bounds to regulate the decomposition.
		In the second stage, the modality-general components are fused
		using weights derived from the predicted margins and distilled
		to the target-modality student.}
	\label{fig:framework}
\end{figure*}

\section{Method}
We propose Transferability-Aware Logit Disentanglement (TALD), a two-stage CMKD framework illustrated in Figure~\ref{fig:framework}.

\subsection{Problem Formulation}
\label{sec:formulation}
In cross-modal knowledge distillation, a teacher $f_T$ pretrained on a source modality and a student $f_S$ trained on a target modality produce logit vectors $\mathbf{z}_T = f_T(x_T) \in \mathbb{R}^K$ and $\mathbf{z}_S = f_S(x_S) \in \mathbb{R}^K$ for a paired sample $x=\{x_T, x_S\}$, where $K$ is the number of categories. Conventional distillation transfers the full teacher logit $\mathbf{z}_T$, which can convey modality-specific cues that conflict with the target modality's discriminative behavior (Figure~\ref{fig:motivation}).

MFH~\cite{mfh} motivates distinguishing modality-general from modality-specific information when considering cross-modal transfer. We extend this perspective to the classification effects of logit components within and across modalities, yielding a behavioral working hypothesis for learning a transferability-aware decomposition.

\textbf{Cross-Modal Transfer Benefit Hypothesis (Benefit Hypothesis).} We represent a modality's logit output $\mathbf{z}$ as the sum of a modality-general component $\mathbf{z}^{gen}$ and a modality-specific component $\mathbf{z}^{spec}$, i.e., $\mathbf{z} = \mathbf{z}^{gen} + \mathbf{z}^{spec}$. The desired classification effects of these components are characterized by two criteria:
\begin{itemize}
	\item \textbf{C1 (Self-benefiting):} The modality-specific component should benefit classification within its own modality: the full logits should yield lower cross-entropy than the modality-general component alone.
	\item \textbf{C2 (Cross-modal exclusivity):} Adding a modality-specific component's cross-modal contribution to the recipient's modality-general logits should increase the recipient's cross-entropy.
\end{itemize}

Here, modality-general and modality-specific describe the intended roles of the learned components. C1 and C2 specify desired classification effects and serve as task-dependent inductive biases for learning the decomposition. A sample-adaptive signed margin tightens or relaxes the required cross-entropy gaps (Sections~\ref{sec:modulator} and~\ref{sec:objective}). In particular, a negative margin permits zero hinge loss even when the corresponding zero-margin criterion is violated.

We introduce a reference student $f_{S_{ref}}$, trained independently on the target modality without distillation. Its logits $\mathbf{z}_{S_{ref}}$ provide the target-modality counterpart to the teacher logits during Stage~1 decomposition, and its learned GEN component contributes to the Stage~2 distillation target.

\subsection{Residual Disentangler}
\label{sec:disentangler}

\begin{figure}[t]
	\centering
	\includegraphics[width=\linewidth]{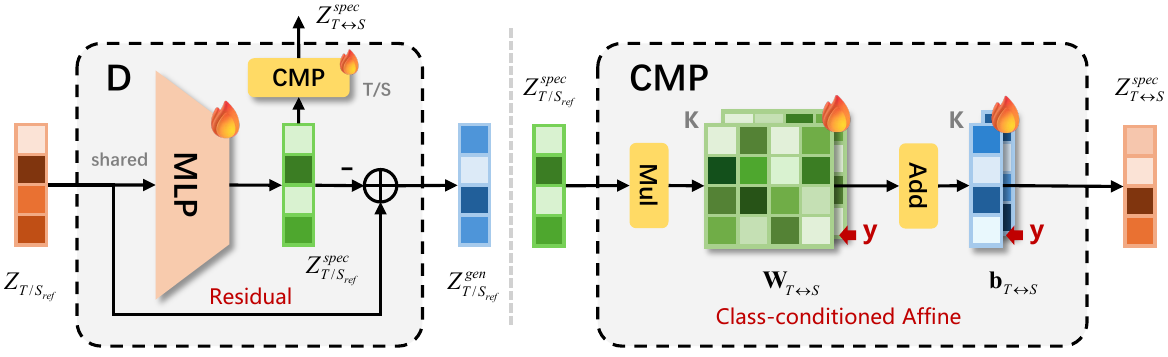}
	\caption{Architecture of the Residual Disentangler. Left: An MLP shared across the two modalities predicts the SPEC components, and the GEN components are obtained by residual subtraction. Right: A direction-specific CMP applies a class-conditioned affine transformation to a SPEC component to construct the cross-modal contribution used in the Stage-1 C2 constraint.}
	\label{fig:disentangler}
\end{figure}

As shown in Figure~\ref{fig:disentangler}, the Residual Disentangler uses a shared three-layer MLP for logit decomposition and direction-specific Cross-Modal Projectors (CMPs) for the C2 constraint. For each logit vector $\mathbf{z}_i \in \mathbb{R}^K$, $i \in \{T, S_{ref}\}$, the MLP predicts the modality-specific component, and the modality-general component is obtained as the residual:

\begin{equation}
	\mathbf{z}_i^{spec} = \mathrm{g}(\mathbf{z}_i),
	\qquad
	\mathbf{z}_i^{gen} = \mathbf{z}_i - \mathbf{z}_i^{spec},
\end{equation}
where $\mathrm{g}(\cdot)$ denotes the MLP shared by the teacher and
reference student. This residual parameterization couples the two
components through their sum, which exactly recovers the original
logits. Since logit differences determine class log-odds, the
modality-specific component acts as an additive correction to the
modality-general class scores.

Each CMP applies a class-conditioned affine transformation to
the source SPEC component (Figure~\ref{fig:disentangler}, right).
The two directions use separate CMP parameters. For the teacher's
component, the transformation is:
\begin{equation}
	\label{eq:cmp_projection}
	\mathbf{z}_{T \to S}^{spec} = \text{P}_{T \to S}(\mathbf{z}_T^{spec}, y) = \mathbf{z}_T^{spec} \cdot \mathbf{W}_{T \to S}^{(y)} + \mathbf{b}_{T \to S}^{(y)},
\end{equation}
where $y$ is the training label,
$\mathbf{W}_{T \to S}^{(y)} \in \mathbb{R}^{K \times K}$ and
$\mathbf{b}_{T \to S}^{(y)} \in \mathbb{R}^{K}$ are class-specific
parameters, and $\mathbf{z}_{T \to S}^{spec}$ is the transformed
component used in C2. The reverse-direction transformation is
defined analogously.

CMP is optimized jointly with the decomposition and used only
in the Stage-1 C2 constraint, where its affine form provides
class- and direction-specific cross-modal contributions.
Stage~2 constructs its distillation target from the GEN
components and the Modulator's fusion weights.
Table~\ref{tab:cmp_ablation} reports downstream design comparisons,
while Appendix~\ref{appendix:cmp} provides frozen-checkpoint
diagnostics of the SPEC-dependent and bias terms.

\subsection{Sample-Adaptive Modulator}
\label{sec:modulator}

\begin{figure}[htbp]
	\centering
	\includegraphics[width=0.7\linewidth]{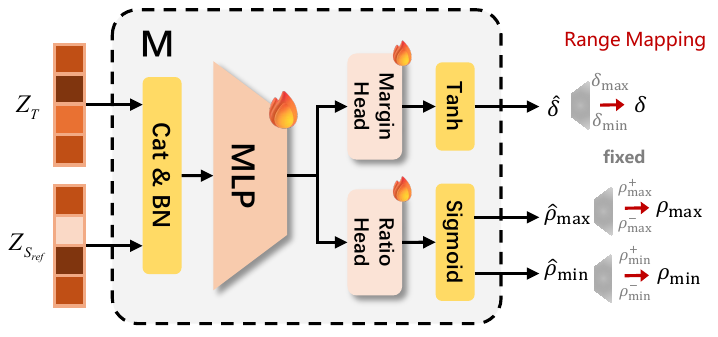}
	\caption{Architecture of the Sample-Adaptive Modulator. A backbone MLP encodes concatenated teacher and reference student logits, from which two independent heads predict margin $\hat{\delta}$ and ratio bounds $[\hat{\rho}_{\min}, \hat{\rho}_{\max}]$, rescaled to predefined intervals via fixed range mapping.}
	\label{fig:modulator}
\end{figure}

As shown in Figure~\ref{fig:modulator}, the Sample-Adaptive Modulator comprises a backbone $\mathrm{MLP}(\cdot)$ and two independent prediction heads. It takes concatenated teacher and reference student logits $\hat{\mathbf{z}} = [\mathbf{z}_T; \mathbf{z}_{S_{ref}}] \in \mathbb{R}^{2K}$ as input, normalized via batch normalization to mitigate scale discrepancies between the two modalities:
\begin{equation}
	\mathbf{h} = \mathrm{MLP}(\text{BN}(\hat{\mathbf{z}})).
\end{equation}

From the hidden representation $\mathbf{h}$, $\text{Head}_{\text{m}}(\cdot)$ predicts a sample-specific signed margin $\delta$ within a predefined interval $[\delta_{\min}, \delta_{\max}]$:
\begin{equation}
	\label{eq:margin_map}
	\delta = \delta_{\min} + \tfrac{1}{2} \bigl( \tanh(\text{Head}_{\text{m}}(\mathbf{h})) + 1 \bigr) (\delta_{\max} - \delta_{\min}).
\end{equation}
The predicted margin $\delta$ adjusts the cross-entropy gaps required by the C1 and C2 hinge losses (Section~\ref{sec:objective}) and is reused to control the adaptive fusion weight (Section~\ref{sec:distillation}).

The second head, $\text{Head}_{\text{r}}(\cdot)$, predicts sample-specific lower and upper bounds on the modality-specific contribution ratio (Section~\ref{sec:objective}). Each bound $\rho_k$ is mapped to a predefined range $[\rho_k^{-}, \rho_k^{+}]$:
\begin{equation}
	\resizebox{0.88\columnwidth}{!}{$
		\rho_k = \rho_k^{-} + \mathrm{sigmoid}(\text{Head}_{\text{r}}(\mathbf{h}))_k
		\left( \rho_k^{+} - \rho_k^{-} \right),
		\quad k \in \{\min, \max\}.$}
\end{equation}

\subsection{Disentanglement Objective}
\label{sec:objective}

The first-stage objective implements the Benefit Hypothesis through transfer benefit constraints, combined with contribution and auxiliary regularization to guide logit decomposition.

\noindent\textbf{Transfer Benefit Constraints.} We implement C1 and C2 as sample-level hinge losses on within- and cross-modal cross-entropy differences. In C2, the cross-modal contribution is evaluated after the CMP transformation:
\begin{equation}
	\resizebox{0.8\columnwidth}{!}{$
		\mathcal{L}_{\text{C1}} = \frac{1}{2}
		\sum_{i \in \{T, S_{ref}\}}
		\mathrm{ReLU}\left(
		\mathrm{CE}(\mathbf{z}_i, y)
		- \mathrm{CE}(\mathbf{z}_i^{gen}, y)
		+ \delta
		\right),
		$}
	\label{eq:L_C1}
\end{equation}
\begin{equation}
	\resizebox{0.9\columnwidth}{!}{$
		\mathcal{L}_{\text{C2}} = \frac{1}{2}
		\sum_{i \in \{T, S_{ref}\}}
		\mathrm{ReLU}\left(
		\mathrm{CE}(\mathbf{z}_i^{gen}, y)
		- \mathrm{CE}(\mathbf{z}_i^{gen}
		+ \mathbf{z}_{\bar{i} \to i}^{spec}, y)
		+ \delta
		\right),
		$}
	\label{eq:L_C2}
\end{equation}
where $\bar{i}$ denotes the other modality and $\delta$ is the sample-adaptive margin predicted by the Modulator (Eq.~\ref{eq:margin_map}). For both hinge losses, positive values of $\delta$ increase the cross-entropy gaps required for zero penalty, while negative values relax these requirements.

The two terms are averaged over a mini-batch of $N$ samples to form the hypothesis loss
$\mathcal{L}_{\text{hyp}} = \frac{1}{2N} \sum_{n=1}^{N}
(\mathcal{L}_{\text{C1}}^{(n)} + \mathcal{L}_{\text{C2}}^{(n)})$.

\noindent\textbf{Relative Contribution Regularization.}
We further regularize the relative classification contribution of the modality-specific component within its own modality. For modality $i \in \{T, S_{ref}\}$, this contribution is measured by the relative cross-entropy reduction obtained by adding the modality-specific component to the modality-general logits:
$\mu_i = \left(
\mathrm{CE}(\mathbf{z}_i^{gen}, y)
- \mathrm{CE}(\mathbf{z}_i, y)
\right) /
\left(
\mathrm{CE}(\mathbf{z}_i^{gen}, y) + \epsilon
\right)$,
where $\epsilon = 10^{-8}$.
Deviations from the sample-specific bounds predicted by the Modulator are penalized through a two-sided hinge loss:
\begin{equation}
	\resizebox{0.85\columnwidth}{!}{$
		\mathcal{R}_{\text{ratio}} = \frac{1}{2}
		\sum_{i \in \{T, S_{ref}\}}
		\Big[
		\mathrm{ReLU}(\rho_{\min} - \mu_i)
		+ \mathrm{ReLU}(\mu_i - \rho_{\max})
		\Big].
		$}
	\label{eq:Rratio}
\end{equation}
Together, $\mathcal{L}_{\text{hyp}}$ guides the within- and cross-modal classification effects of the decomposition, while $\mathcal{R}_{\text{ratio}}$ regularizes the relative contribution of each modality-specific component within its own modality.

\noindent\textbf{Auxiliary Regularization.}
We apply a supervised contrastive loss $\mathcal{L}_{\text{con}}$~\cite{scl} to the hidden representation $\mathbf{h}$ to encourage same-class samples to cluster, and a variance-band regularizer $\mathcal{L}_{\mathrm{var}}$ to control the dispersion of the sample-specific margins. These terms form the auxiliary regularization
$\mathcal{R}_{\text{aux}} =
\lambda_{\text{con}} \mathcal{L}_{\text{con}}
+ \lambda_{\text{var}} \mathcal{L}_{\text{var}}$,
with detailed formulations provided in Appendix~\ref{appendix:aux_reg}.

\noindent\textbf{Total Objective.}
The first-stage objective is:
\begin{equation}
	\mathcal{L}_{LD}
	= \lambda_{\text{hyp}} \mathcal{L}_{\text{hyp}}
	+ \lambda_{\text{ratio}} \mathcal{R}_{\text{ratio}}
	+ \lambda_{\text{aux}} \mathcal{R}_{\text{aux}},
\end{equation}
where $\lambda_{\text{hyp}}$, $\lambda_{\text{ratio}}$, and $\lambda_{\text{aux}}$ are weighting coefficients. Stage~1 training uses early stopping based on validation loss.

\begin{table*}[t]
	\caption{Performance comparison on five multimodal classification
		benchmarks. Accuracy is averaged over five runs.
		$\Delta$Acc is the mean gain (pp) over w/o KD across ten
		transfer directions. PTF is the percentage of the 50 evaluations
		(ten directions, five runs each) with strictly positive accuracy
		gains over the corresponding w/o KD baseline.
		The best and second-best results per column are \textbf{bold},
		with the best additionally \textbf{\underline{underlined}}.}
	\label{tab:cm_performance}
	\centering
	{%
		\fontsize{9}{11}\selectfont
		\rmfamily
		\renewcommand{\arraystretch}{0.85}
		\setlength{\tabcolsep}{0.85mm}

		\edef\CompactHeaderShift{%
			\the\dimexpr0pt-(0.4ex+\cmidrulewidth)/2\relax}

		\begin{tabular}{@{}ll*{12}{c}@{}}
			\toprule
			\multirow[c]{2}{*}[\CompactHeaderShift]{\textbf{Paradigm}}
			& \multirow[c]{2}{*}[\CompactHeaderShift]{\textbf{Method}}
			& \multicolumn{2}{c}{\textbf{AV-MNIST}}
			& \multicolumn{2}{c}{\textbf{RAVDESS}}
			& \multicolumn{2}{c}{\textbf{VGGSound-50k}}
			& \multicolumn{2}{c}{\textbf{CrisisMMD-V2}}
			& \multicolumn{2}{c}{\textbf{Fakeddit}}
			& \multirow[c]{2}{*}[\CompactHeaderShift]{\textbf{$\Delta$Acc}}
			& \multirow[c]{2}{*}[\CompactHeaderShift]{\textbf{PTF}} \\
			\noalign{\vskip\dimexpr0.2ex-\aboverulesep\relax}
			\cmidrule(lr){3-4}
			\cmidrule(lr){5-6}
			\cmidrule(lr){7-8}
			\cmidrule(lr){9-10}
			\cmidrule(lr){11-12}
			\noalign{\vskip\dimexpr0.2ex-\belowrulesep\relax}
			& &
			Image & Audio &
			Visual & Audio &
			Visual & Audio &
			Image & Text &
			Image & Text & & \\
			\midrule

			\textemdash & w/o KD
			& 0.6322 & 0.4372 & 0.7472 & 0.6694
			& 0.4080 & 0.5818 & 0.5524 & 0.5405
			& 0.5729 & 0.5196 & \textemdash & \textemdash \\
			\hdashline
			\noalign{\vskip 2pt}

			\multirow{2}{*}{Response-based} & KD
			& 0.6328 & 0.4377 & 0.7035 & 0.6944
			& 0.3855 & 0.5887 & 0.5501 & 0.5377
			& 0.5745 & 0.4999 & -0.56 & 42\% \\

			& MLLD
			& 0.6327 & 0.4380 & 0.7208 & 0.6799
			& 0.4044 & 0.5898 & 0.5541 & 0.5127
			& 0.5726 & 0.2610 & -2.95 & 46\% \\
			\midrule

			\multirow{2}{*}{Feature-based} & FitNets
			& 0.6201 & 0.4292 & 0.6889 & 0.6229
			& 0.3742 & 0.5685 & 0.5458 & 0.5165
			& 0.5492 & \textbf{\underline{0.5465}}
			& -1.99 & 10\% \\

			& OFA
			& \textbf{0.6334} & 0.4380 & 0.7292 & 0.6445
			& 0.3861 & 0.5839 & 0.5493 & \textbf{0.5420}
			& 0.5727 & \textbf{0.5330} & -0.49 & 50\% \\
			\midrule

			\multirow{2}{*}{Relation-based} & RKD
			& \textbf{\underline{0.6337}} & 0.3850
			& 0.7569 & 0.6826 & 0.4092 & 0.5709
			& 0.5508 & 0.5419 & 0.5737 & 0.1681
			& -3.88 & 42\% \\

			& CRD
			& 0.6281 & \textbf{0.4389} & 0.7736 & 0.7014
			& 0.3537 & 0.5826 & 0.5504 & 0.5256
			& 0.5741 & 0.3706 & -1.62 & 38\% \\
			\midrule

			\multirow{4}{*}{Cross-Modal} & MGDFR
			& 0.6329 & \textbf{0.4389}
			& 0.7576 & \textbf{0.7111}
			& 0.4375 & \textbf{0.5967}
			& 0.5469 & 0.5390
			& \textbf{0.5766} & 0.5239
			& \textbf{+1.00} & \textbf{70\%} \\

			& C\textsuperscript{2}KD
			& 0.6309 & \textbf{0.4389}
			& \textbf{\underline{0.7754}} & 0.6500
			& 0.4229 & \textbf{\underline{0.6080}}
			& 0.5528 & 0.5204 & 0.2897 & 0.1856
			& -5.87 & 44\% \\

			& FD-CMKD
			& 0.6261 & \textbf{\underline{0.4396}}
			& 0.7604 & 0.6896
			& \textbf{0.4437} & 0.5749
			& \textbf{0.5545} & 0.5388
			& 0.5729 & 0.5168 & +0.56 & 46\% \\

			\rowcolor{gray!10}
			\cellcolor{white} & \textbf{TALD}
			& 0.6328 & 0.4383
			& \textbf{0.7750} & \textbf{\underline{0.7146}}
			& \textbf{\underline{0.4484}} & 0.5939
			& \textbf{\underline{0.5567}}
			& \textbf{\underline{0.5427}}
			& \textbf{\underline{0.5772}} & 0.5194
			& \textbf{\underline{+1.38}}
			& \textbf{\underline{84\%}} \\
			\bottomrule
		\end{tabular}
	}
\end{table*}

\subsection{Adaptive Knowledge Distillation}
\label{sec:distillation}
After Stage~1, we freeze the Disentangler and Modulator. Stage~2 trains the student $f_S$ using distillation targets constructed from the learned decomposition and adaptive fusion weights.

\noindent\textbf{Adaptive Distillation Target.} The Modulator-predicted margin $\delta$ (Eq.~\ref{eq:margin_map}) is reused as a learned control signal for sample-wise fusion of the two GEN sources. We scale it by a sensitivity coefficient $\beta$ and apply a sigmoid to obtain the teacher fusion weight:
\begin{equation}
	w = \mathrm{sigmoid}(\beta \cdot \delta),
	\label{eq:adaptive_weight}
\end{equation}
where $\beta$ controls the dynamic range of $w$ when the margin interval $[\delta_{\min}, \delta_{\max}]$ is narrow. The target then fuses the modality-general components of the teacher and reference student:
\begin{equation}
	\mathbf{z}_{target} = w \cdot \mathbf{z}_T^{gen} + (1 - w) \cdot \mathbf{z}_{S_{ref}}^{gen}.
\end{equation}
\noindent\textbf{Distillation Loss.} The student is trained with a standard distillation objective~\cite{kd}:
\begin{equation}
	\mathcal{L} = (1 - \alpha) \cdot \mathrm{CE}(\mathbf{z}_S, y) + \alpha \cdot T^2 \cdot \mathrm{KL}( \mathbf{q}_{target} \| \mathbf{q}_S),
\end{equation}
where $\mathbf{q}_{target} = \mathrm{softmax}(\mathbf{z}_{target} / T)$, $\mathbf{q}_S = \mathrm{softmax}(\mathbf{z}_S / T)$, $\alpha$ is the distillation weight, $\mathrm{KL}(\cdot \| \cdot)$ is the KL-divergence, and $T$ is the temperature.

\section{Experiments}
We evaluate TALD on five cross-modal classification benchmarks spanning image, video, audio, and text: AV-MNIST~\cite{avmnist}, RAVDESS~\cite{ravdess}, VGGSound-50k~\cite{vggs}, CrisisMMD-V2~\cite{cmmdv2}, and Fakeddit~\cite{fakeddit}. We compare against nine baselines grouped into four categories: response-based (KD~\cite{kd}, MLLD~\cite{mlld}), feature-based (FitNets~\cite{fitnets}, OFA~\cite{ofa}), relation-based (RKD~\cite{rkd}, CRD~\cite{crd}), and cross-modal (MGDFR~\cite{mfh}, C\textsuperscript{2}KD~\cite{c2kd}, FD-CMKD~\cite{fd-cmkd}). We assess accuracy and computational efficiency, examine target construction and training design through ablations, and visualize the learned logit components and sample distributions to analyze transfer behavior.

\subsection{Implementation Details}
We retrain all methods on identically preprocessed data over five runs with different random seeds and 60\%/20\%/20\% train/validation/test splits. Corresponding splits are shared across methods. Within each dataset, methods use the same teacher and student architectures, common hyperparameter settings, and 100 distillation epochs. Method-specific hyperparameters follow the original papers or are selected using the validation set. We report mean test accuracy from validation-selected checkpoints. TALD uses $\beta=5$ and $[\delta_{\min},\delta_{\max}]=[-0.1,0.1]$, with up to 30 Stage~1 epochs and early stopping based on validation loss. Dataset preparation, model configurations, and training settings are detailed in Appendix~\ref{appendix:moreI&E}.

\subsection{Comparison with State-of-the-Art}
\paragraph{Performance Comparison.}
Table~\ref{tab:cm_performance} shows that CMKD gains vary across datasets and transfer directions. TALD achieves the largest mean accuracy gain over w/o KD (+1.38 pp), ranks first in five of ten directions, and attains the highest overall positive-transfer frequency (84\%). Its modest gains on image--audio and image--text benchmarks may partly reflect weaker task-relevant correspondence than in video--audio benchmarks~\cite{mfh}. Further analyses of gain magnitudes and per-direction positive-transfer frequencies are provided in Appendix~\ref{appendix:positive_transfer_frequency}.

\begin{table}[t]
	\caption{Efficiency comparison of four CMKD methods. Stage costs are averaged over five datasets and both directions. S0 denotes unimodal pretraining; the remaining stages follow each method's pipeline. Total sums trainable parameters (Params), computation (FLOPs), and wall-clock time (Time), and takes the maximum of the stage-averaged peak GPU memory (Mem).}
	\label{tab:cost_e2e}
	\centering
	{%
		\fontsize{9}{11}\selectfont
		\rmfamily
		\renewcommand{\arraystretch}{0.7}
		\setlength{\tabcolsep}{0.8mm}

		\begin{tabular}{@{}llrrrrc@{}}
			\toprule
			\multirow{2}{*}{\textbf{Method}}
			& \multirow{2}{*}{\textbf{Stage}}
			& \multicolumn{1}{c}{\textbf{Params}}
			& \multicolumn{1}{c}{\textbf{FLOPs}}
			& \multicolumn{1}{c}{\textbf{Mem}}
			& \multicolumn{1}{c}{\textbf{Time}}
			& \multicolumn{1}{c}{\textbf{$\Delta$Acc}} \\
			&
			& \multicolumn{1}{c}{(M) $\downarrow$}
			& \multicolumn{1}{c}{(T) $\downarrow$}
			& \multicolumn{1}{c}{(MiB) $\downarrow$}
			& \multicolumn{1}{c}{(s) $\downarrow$}
			& \multicolumn{1}{c}{(pp) $\uparrow$} \\
			\midrule

			\multirow{4}{*}{MGDFR}
			& S1 & 3.98 & 4871.12 & 1536.39 & 872.96
			& \multirow{4}{*}{+1.00} \\
			& S2 & 0.00 & 18260.23 & 863.03 & 814.39 & \\
			& S3 & 1.99 & 3589.71 & 2406.10 & 874.02 & \\
			& \cellcolor{gray!10}\textbf{Total}
			& \cellcolor{gray!10}5.96
			& \cellcolor{gray!10}26721.05
			& \cellcolor{gray!10}2406.10
			& \cellcolor{gray!10}2561.36 & \\
			\midrule

			\multirow{3}{*}{C\textsuperscript{2}KD}
			& S0 & 1.99 & 2435.64 & 1530.85 & 653.05
			& \multirow{3}{*}{-5.87} \\
			& S1 & 4.30 & 4874.97 & 1554.16 & 894.54 & \\
			& \cellcolor{gray!10}\textbf{Total}
			& \cellcolor{gray!10}6.28
			& \cellcolor{gray!10}7310.61
			& \cellcolor{gray!10}1554.16
			& \cellcolor{gray!10}1547.59 & \\
			\midrule

			\multirow{3}{*}{FD-CMKD}
			& S0 & 1.99 & 2435.64 & 1530.85 & 653.05
			& \multirow{3}{*}{+0.56} \\
			& S1 & 65.48 & 4666.04 & 2515.24 & 878.85 & \\
			& \cellcolor{gray!10}\textbf{Total}
			& \cellcolor{gray!10}67.47
			& \cellcolor{gray!10}7101.68
			& \cellcolor{gray!10}2515.24
			& \cellcolor{gray!10}1531.91 & \\
			\midrule

			\multirow{4}{*}{\textbf{TALD}}
			& S0 & 1.99 & 2435.64 & 1530.85 & 653.05
			& \multirow{4}{*}{+1.38} \\
			& S1 & 1.15 & 692.76 & 1097.27 & 271.37 & \\
			& S2 & 1.99 & 4743.86 & 1718.08 & 915.94 & \\
			& \cellcolor{gray!10}\textbf{Total}
			& \cellcolor{gray!10}5.13
			& \cellcolor{gray!10}7872.26
			& \cellcolor{gray!10}1718.08
			& \cellcolor{gray!10}1840.36 & \\
			\bottomrule
		\end{tabular}
	}
\end{table}

\paragraph{Efficiency Comparison.}
Table~\ref{tab:cost_e2e} shows that TALD reduces total FLOPs, training time, and peak GPU memory by 70.5\%, 28.1\%, and 28.6\%, respectively, relative to MGDFR. Its disentanglement stage accounts for 8.8\% of total FLOPs and 14.7\% of training time. Although C$^2$KD and FD-CMKD require fewer FLOPs and less training time, TALD achieves the largest mean accuracy gain, offering a favorable accuracy--efficiency trade-off. Detailed costs and accuracy gains are provided in Appendix~\ref{appendix:efficiency}.

\begin{table}[t]
	\caption{Framework design ablation. $T$ and $S_{\mathrm{ref}}$ denote the cross-modal teacher and independently trained target-modality reference student. Each dataset score averages both directions; Avg averages the two datasets.}
	\label{tab:framework}
	\centering
	{%
		\fontsize{9}{11}\selectfont
		\rmfamily
		\renewcommand{\arraystretch}{0.85}
		\setlength{\tabcolsep}{0.55mm}

		\begin{tabular}{@{}clcccc@{}}
			\toprule
			\textbf{\#}
			& \textbf{Framework}
			& \textbf{Sources}
			& \textbf{RAVDESS}
			& \textbf{Fakeddit}
			& \textbf{Avg} \\
			\midrule

			(a)
			& w/o KD
			& \textemdash
			& 0.7083
			& \textbf{0.5463}
			& 0.6273 \\
			\midrule

			(b)
			& Vanilla KD
			& $T$
			& 0.6990
			& 0.5372
			& 0.6181 \\

			(c)
			& Vanilla KD
			& $S_{\mathrm{ref}}$
			& 0.7260
			& 0.5452
			& 0.6356 \\

			(d)
			& Mean Fusion
			& $T+S_{\mathrm{ref}}$
			& 0.7250
			& 0.5439
			& 0.6345 \\

			(e)
			& Single Stage
			& $T+S_{\mathrm{ref}}$
			& \textbf{0.7365}
			& 0.5443
			& \textbf{0.6404} \\

			\rowcolor{gray!10}
			(f)
			& \textbf{Two Stage}
			& $T+S_{\mathrm{ref}}$
			& \textbf{\underline{0.7448}}
			& \textbf{\underline{0.5483}}
			& \textbf{\underline{0.6466}} \\
			\bottomrule
		\end{tabular}
	}
\end{table}

\subsection{Ablation Studies}
We examine TALD's design choices on RAVDESS and Fakeddit, covering video--audio and image--text transfer. The ablations first assess supervision sources and training stages, then examine target construction, module configurations, and learning objectives. We additionally evaluate hyperparameter sensitivity on RAVDESS.

\paragraph{Framework Design.}
Table~\ref{tab:framework} examines the supervision sources and training stages. Teacher-only KD (b) underperforms w/o KD on both datasets, while reference-only KD (c) improves on (b) but remains below the baseline on Fakeddit. Mean fusion (d) underperforms (c), indicating that simply combining raw responses is insufficient for consistent gains. Single-stage TALD (e) improves over mean fusion; separating decomposition from distillation (f) adds 0.83 and 0.40 percentage points on RAVDESS and Fakeddit, respectively, yielding the highest mean accuracy on both datasets. These results support a target-modality reference together with learned target construction and a two-stage training pipeline.

\paragraph{Distillation Target Design.}
Within the two-stage framework, Table~\ref{tab:target_ablation} examines component selection and fusion using both the teacher and reference student. Under adaptive fusion, GEN outperforms ORG and SPEC on both datasets; SPEC is RMS-matched to GEN to reduce the influence of logit scale. With GEN fixed, adaptive fusion also improves over equal weighting. Shuffling the learned weights across samples (Shuffled-S) or within classes (Shuffled-C) reduces Avg by 0.66 and 0.56 percentage points, respectively. Both controls preserve the overall weight distribution, and Shuffled-C additionally preserves each class's distribution. The resulting drops support the value of sample--weight correspondence beyond global or class-level weighting. Together, these comparisons support selecting GEN and adapting its fusion to individual samples. Complete source comparisons and SPEC scaling details are provided in Appendix~\ref{appendix:target_ablation}.

\begin{table}[t]
	\caption{Ablation of logit selection and adaptive fusion
		using both the teacher and reference student.
		ORG, GEN, and SPEC denote original, modality-general,
		and modality-specific logits. Mean and Adaptive use
		equal weights and the learned weight $w$, respectively.
		Shuffled-S permutes $w$ across samples;
		Shuffled-C restricts the permutation to each class.}
	\label{tab:target_ablation}
	\centering
	{%
		\fontsize{9}{11}\selectfont
		\rmfamily
		\renewcommand{\arraystretch}{0.85}
		\setlength{\tabcolsep}{0.85mm}
		\begin{tabular}{@{}cccccc@{}}
			\toprule
			\textbf{\#} & \textbf{Logits} & \textbf{Fusion} & \textbf{RAVDESS} & \textbf{Fakeddit} & \textbf{Avg} \\
			\midrule
			(a) & ORG & Adaptive
			& \textbf{0.7410} & 0.5443 & \textbf{0.6427} \\
			(b) & SPEC & Adaptive
			& 0.7247 & 0.5428 & 0.6338 \\
			\midrule
			(c) & GEN & Mean
			& 0.7358 & 0.5430 & 0.6394 \\
			(d) & GEN & Shuffled-S
			& 0.7368 & 0.5431 & 0.6400 \\
			(e) & GEN & Shuffled-C
			& 0.7375 & \textbf{0.5445} & 0.6410 \\
			\rowcolor{gray!10}
			\textbf{(f)} & GEN & Adaptive
			& \textbf{\underline{0.7448}} & \textbf{\underline{0.5483}} & \textbf{\underline{0.6466}} \\
			\bottomrule
		\end{tabular}
	}
\end{table}

\begin{table}[t]
	\caption{Module design ablation. IND, FS, and PS denote independent, fully shared, and partially shared disentanglers; PS shares the MLP and uses separate CMPs. ResDis and Adapt denote residual disentanglement and sample-adaptive modulation.}
	\label{tab:module}
	\centering
	{%
		\fontsize{9}{11}\selectfont
		\rmfamily
		\renewcommand{\arraystretch}{0.85}
		\setlength{\tabcolsep}{0.55mm}

		\edef\CompactHeaderShift{%
			\the\dimexpr0pt-(0.4ex+\cmidrulewidth)/2\relax}
		\begin{tabular}{@{}ccccccc@{}}
			\toprule
			\multirow[c]{2}{*}[\CompactHeaderShift]{\textbf{\#}}
			& \multicolumn{2}{c}{\textbf{Disentangler}}
			& \textbf{Modulator}
			& \multirow[c]{2}{*}[\CompactHeaderShift]{\textbf{RAVDESS}}
			& \multirow[c]{2}{*}[\CompactHeaderShift]{\textbf{Fakeddit}}
			& \multirow[c]{2}{*}[\CompactHeaderShift]{\textbf{Avg}} \\
			\noalign{\vskip\dimexpr0.2ex-\aboverulesep\relax}
			\cmidrule(lr){2-3}
			\cmidrule(lr){4-4}
			\noalign{\vskip\dimexpr0.2ex-\belowrulesep\relax}
			& Sharing
			& ResDis
			& Adapt
			& & & \\
			\midrule

			(a)
			& IND
			& \ding{55}
			& \ding{55}
			& 0.7400
			& 0.5481
			& 0.6441 \\

			(b)
			& IND
			& \ding{51}
			& \ding{51}
			& 0.7351
			& 0.5460
			& 0.6406 \\
			\midrule

			(c)
			& FS
			& \ding{55}
			& \ding{55}
			& 0.7385
			& \textbf{\underline{0.5492}}
			& 0.6439 \\

			(d)
			& FS
			& \ding{51}
			& \ding{51}
			& \textbf{0.7413}
			& 0.5480
			& \textbf{0.6447} \\
			\midrule

			(e)
			& PS
			& \ding{55}
			& \ding{55}
			& 0.7337
			& 0.5474
			& 0.6406 \\

			(f)
			& PS
			& \ding{55}
			& \ding{51}
			& 0.7278
			& 0.5481
			& 0.6380 \\

			(g)
			& PS
			& \ding{51}
			& \ding{55}
			& 0.7403
			& 0.5455
			& 0.6429 \\

			\rowcolor{gray!10}
			\textbf{(h)}
			& PS
			& \ding{51}
			& \ding{51}
			& \textbf{\underline{0.7448}}
			& \textbf{0.5483}
			& \textbf{\underline{0.6466}} \\
			\bottomrule
		\end{tabular}
	}
\end{table}

\paragraph{Module Design.}
Table~\ref{tab:module} examines the modules implementing this target construction. With ResDis and Adapt enabled, PS (h) outperforms IND (b) and FS (d) on both datasets, supporting shared decomposition with direction-specific CMPs. Within PS, their joint use performs best, despite mixed effects from either alone. The full configuration achieves the highest Avg, although FS (c) performs best on Fakeddit. These results support combining the residual coupling of GEN and SPEC with sample-adaptive constraints and fusion.

\begin{figure}[t]
	\centering
	\includegraphics[width=0.95\linewidth]{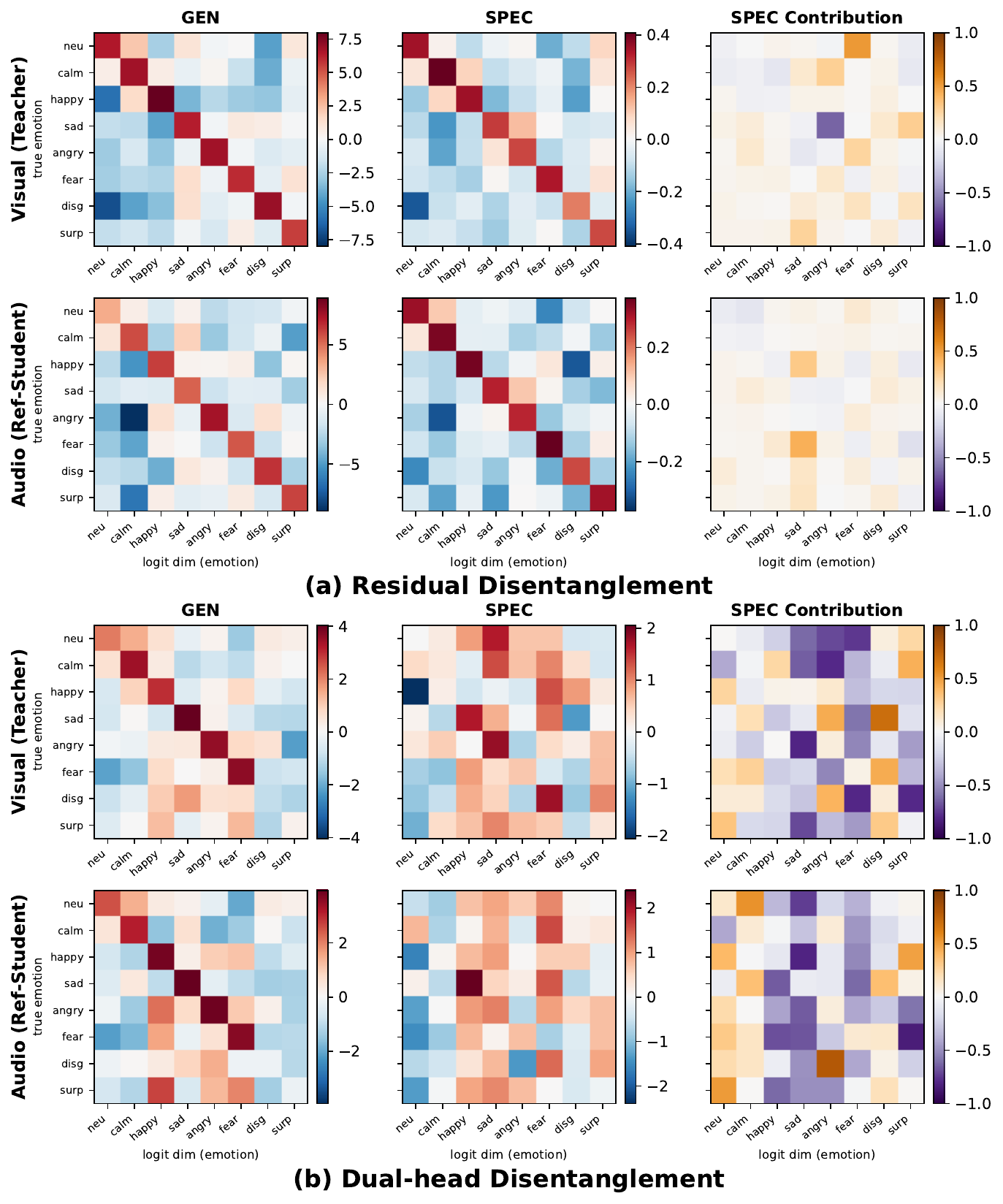}
	\caption{Per-class heatmaps of the learned logit components. (a) Residual disentanglement (ours): GEN retains a clear class-discriminative diagonal, SPEC exhibits class-dependent corrections with near-zero mean, and the SPEC contribution $-z^{spec}/(|z^{org}|+|z^{spec}|)$ is strongly negative at (true \emph{Sad}, pred.\ \emph{Angry}). (b) The dual-head variant produces larger SPEC values, comparable in magnitude to GEN, with less distinct class structure in the displayed heatmaps.}
	\label{fig:heatmaps}
\end{figure}

The class-independent affine CMP (b) in Table~\ref{tab:cmp_ablation} already outperforms w/o KD on both datasets. The class-conditioned affine mapping (e) further raises Avg from 0.6434 to 0.6466, supporting class-dependent modeling of cross-modal contributions. Its advantage over bias-only (d) also supports retaining SPEC-dependent projection. The SPEC heatmaps in Figure~\ref{fig:heatmaps} provide complementary qualitative evidence of class-dependent discriminative structure. Further diagnostics of the SPEC-dependent and bias terms are provided in Appendix~\ref{appendix:cmp}.

\begin{table}[t]
	\caption{CMP design ablation. ID, CI, and CC denote identity,
		class-independent, and class-conditioned mappings, respectively.
		ID uses the other modality's SPEC directly in C2.
		Linear omits the bias; Bias-only omits the SPEC-dependent term.}
	\label{tab:cmp_ablation}
	\centering
	{%
		\fontsize{9}{11}\selectfont
		\rmfamily
		\renewcommand{\arraystretch}{1.0}
		\setlength{\tabcolsep}{3pt}
		\begin{tabular}{@{}clcccc@{}}
			\toprule
			\textbf{\#}
			& \textbf{Setting}
			& \textbf{Mapping}
			& \textbf{RAVDESS}
			& \textbf{Fakeddit}
			& \textbf{Avg} \\
			\midrule
			(a) & w/o CMP & ID
			& \textbf{0.7441}
			& 0.5437
			& 0.6439 \\
			(b) & Affine & CI
			& 0.7396
			& \textbf{0.5472}
			& 0.6434 \\
			(c) & Linear & CC
			& 0.7439
			& 0.5461
			& \textbf{0.6450} \\
			(d) & Bias-only & CC
			& 0.7425
			& 0.5434
			& 0.6430 \\
			\rowcolor{gray!10}
			(e) & \textbf{Affine} & \textbf{CC}
			& \textbf{\underline{0.7448}}
			& \textbf{\underline{0.5483}}
			& \textbf{\underline{0.6466}} \\
			\bottomrule
		\end{tabular}
	}
\end{table}

\paragraph{Loss Configuration.}
Table~\ref{tab:loss_ablation} shows that C1 (d) improves over noHyp (c) with both regularizers retained. Adding C2 (g) yields further gains of 0.38 and 0.05 percentage points on RAVDESS and Fakeddit, respectively. Without regularization, the joint criteria (e) underperform either alone (a,b) in Avg. Adding $\mathcal{R}_{\mathrm{ratio}}$ (f) and $\mathcal{R}_{\mathrm{aux}}$ (g) successively improves both datasets. These results support complementing residual reconstruction with joint constraints on component contributions and adaptive modulation. Individual auxiliary terms are examined in Appendix~\ref{appendix:aux_reg}.

\begin{table}[t]
	\caption{Loss configuration ablation. $\mathcal{L}_{\mathrm{C1}}$ and $\mathcal{L}_{\mathrm{C2}}$ implement the self-benefiting and cross-modal exclusivity criteria; $\mathcal{R}_{\mathrm{ratio}}$ and $\mathcal{R}_{\mathrm{aux}}$ denote contribution and auxiliary regularization. Rows (c) and (d) denote noHyp and w/o C2, respectively. Rows (e)--(g) progressively add regularization to the joint criterion loss.}
	\label{tab:loss_ablation}
	\centering
	{%
		\fontsize{9}{11}\selectfont
		\rmfamily
		\renewcommand{\arraystretch}{0.85}
		\setlength{\tabcolsep}{0.55mm}

		\edef\CompactHeaderShift{%
			\the\dimexpr0pt-(0.4ex+\cmidrulewidth)/2\relax}

		\begin{tabular}{@{}cccccccc@{}}
			\toprule
			\multirow[c]{2}{*}[\CompactHeaderShift]{\textbf{\#}}
			& \multicolumn{2}{c}{$\mathcal{L}_{\mathrm{hyp}}$}
			& \multirow[c]{2}{*}[\CompactHeaderShift]{$\mathcal{R}_{\mathrm{ratio}}$}
			& \multirow[c]{2}{*}[\CompactHeaderShift]{$\mathcal{R}_{\mathrm{aux}}$}
			& \multirow[c]{2}{*}[\CompactHeaderShift]{\textbf{RAVDESS}}
			& \multirow[c]{2}{*}[\CompactHeaderShift]{\textbf{Fakeddit}}
			& \multirow[c]{2}{*}[\CompactHeaderShift]{\textbf{Avg}} \\

			\noalign{\vskip\dimexpr0.2ex-\aboverulesep\relax}
			\cmidrule(lr){2-3}
			\noalign{\vskip\dimexpr0.2ex-\belowrulesep\relax}

			& $\mathcal{L}_{\mathrm{C1}}$
			& $\mathcal{L}_{\mathrm{C2}}$
			& & & & & \\
			\midrule

			(a)
			& \ding{51} & \ding{55} & \ding{55} & \ding{55}
			& 0.7403 & 0.5448 & 0.6426 \\

			(b)
			& \ding{55} & \ding{51} & \ding{55} & \ding{55}
			& 0.7399 & 0.5456 & 0.6428 \\

			(c)
			& \ding{55} & \ding{55} & \ding{51} & \ding{51}
			& 0.7393 & 0.5452 & 0.6423 \\

			(d)
			& \ding{51} & \ding{55} & \ding{51} & \ding{51}
			& 0.7410 & \textbf{0.5478} & \textbf{0.6444} \\

			\midrule

			(e)
			& \ding{51} & \ding{51} & \ding{55} & \ding{55}
			& 0.7369 & 0.5453 & 0.6411 \\

			(f)
			& \ding{51} & \ding{51} & \ding{51} & \ding{55}
			& \textbf{0.7421} & 0.5466 & \textbf{0.6444} \\

			\rowcolor{gray!10}
			\textbf{(g)}
			& \ding{51} & \ding{51} & \ding{51} & \ding{51}
			& \textbf{\underline{0.7448}}
			& \textbf{\underline{0.5483}}
			& \textbf{\underline{0.6466}} \\

			\bottomrule
		\end{tabular}
	}
\end{table}

\begin{figure}[t]
	\centering
	\includegraphics[width=\linewidth]{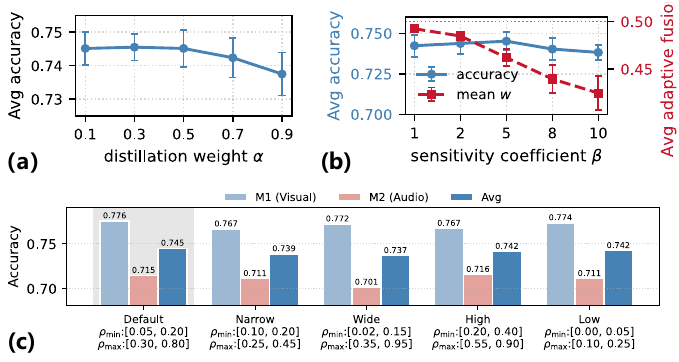}
	\caption{Hyperparameter sensitivity on RAVDESS:
		(a) distillation weight $\alpha$;
		(b) sensitivity coefficient $\beta$ and mean fusion weight $w$;
		(c) output ranges of $\rho_{\min}$ and $\rho_{\max}$.
		Results are averaged over five runs; error bars in (a) and (b)
		indicate standard deviations.}
	\label{fig:sensitivity}
\end{figure}

\paragraph{Hyperparameter Sensitivity.}
Figure~\ref{fig:sensitivity} shows relatively stable accuracy at moderate $\alpha$ and $\beta$ on RAVDESS. Increasing $\beta$ lowers the mean teacher weight and amplifies sample-wise weight differences within the tested range (Eq.~\ref{eq:adaptive_weight}), but accuracy declines slightly at larger values, favoring moderate sensitivity. We use an initial distillation weight of $\alpha=0.5$ and set $\beta=5$. For the ratio bounds, $[\rho_{\min}^{-},\rho_{\min}^{+}]=[0.05,0.2]$ and $[\rho_{\max}^{-},\rho_{\max}^{+}]=[0.3,0.8]$ yield the highest tested mean accuracy and are adopted as defaults. These settings are shared across all five benchmarks, on which TALD achieves positive mean gains (Table~\ref{tab:cm_performance}).

\begin{figure}[t]
	\centering
	\includegraphics[width=1\columnwidth]{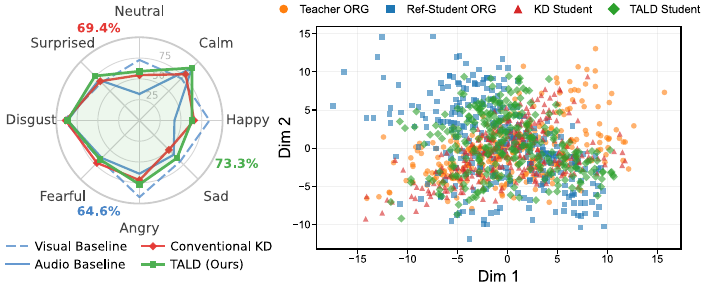}
	\caption{Class-wise performance and projected logit distributions from a single RAVDESS V$\to$A run.
		Left: per-class accuracy of the visual and audio baselines, the
		KD student, and TALD, where KD falls below the audio baseline on
		\emph{Sad} and \emph{Calm} while TALD stays above it on all classes.
		Right: in the displayed PCA projection, the KD student overlaps
		with the teacher's source-modality cluster, while TALD lies
		closer to the target-modality cluster.}
	\label{fig:dist_collapse}
\end{figure}

\subsection{Visualization and Analysis}
\label{sec:semantic_validity}

\paragraph{Class-level structure of residual decomposition.}
The residual formulation expresses logits as a GEN base plus a SPEC correction, coupling both components through exact reconstruction. To examine whether this design preserves class-discriminative structure with modest corrections, Figure~\ref{fig:heatmaps} compares residual and dual-head decompositions using class-wise heatmaps. Residual GEN retains a clear diagonal, while SPEC exhibits class-dependent responses at much smaller magnitudes. For \emph{Sad} samples, subtracting SPEC lowers the visual teacher's competing \emph{Angry} score. The dual-head variant produces larger SPEC values and more diffuse corrections. These observations support residual coupling as an inductive bias for preserving emotion-discriminative structure while learning selective adjustments to class responses.

\paragraph{Sample distribution and transfer behavior.}
We next examine whether the resulting supervision preserves target-modality structure across samples while improving transfer behavior (Figure~\ref{fig:dist_collapse}). In a single RAVDESS V$\to$A run, TALD appears closer to the audio reference in the PCA projection, while conventional KD overlaps more with the visual teacher. The radar plot complements this structural observation: KD falls below the audio baseline on \emph{Sad} and \emph{Calm} despite gains on other classes, whereas TALD remains above it across all displayed classes. These patterns are consistent with retaining target-modality discrimination while mitigating class-dependent negative transfer. Additional component visualizations are provided in Appendix~\ref{appendix:disentangle_vis}.

\section{Discussion}
\label{sec:criteria_status}

TALD extends the MFH perspective~\cite{mfh} through an idealized separation of modality-general and modality-specific knowledge. The Benefit Hypothesis operationalizes this separation through within- and cross-modal classification effects, making transferability relative to the recipient modality and task. Residual coupling anchors the components to the original predictions, while adaptive regulation and class-conditioned CMPs accommodate sample and category variation. The fusion and loss ablations support selectively combining cross-modal evidence with target-modality discrimination (Tables~\ref{tab:target_ablation} and~\ref{tab:loss_ablation}); the sensitivity analysis further suggests balancing teacher contribution and sample-wise weight variation (Figure~\ref{fig:sensitivity}(b)). Training dynamics show early reductions in both zero-margin criterion losses, with subsequent satisfaction and relaxation varying across transfer directions (Appendices~\ref{appendix:margin_range} and~\ref{appendix:full_dynamics}). These patterns are consistent with asymmetric transferability, but also depend on prediction confidence, learned projections, and margin regulation. The criteria thus guide useful target construction without establishing strict semantic separation (Appendix~\ref{appendix:cmp}). Future work could incorporate category-level inter-modal agreement to develop decomposition constraints better suited to specific tasks and transfer directions.

\section{Conclusion}

We presented TALD, a two-stage framework that translates the MFH perspective into task-dependent self-benefiting and cross-modal exclusivity criteria for residual logit decomposition. Adaptive fusion combines the teacher's and reference student's learned modality-general components to construct supervision that accounts for both modalities. Across five benchmarks and ten transfer directions, TALD achieves a mean accuracy gain of 1.38 percentage points over training without distillation, with a favorable accuracy--efficiency trade-off. Ablations and visualizations support the utility of the proposed decomposition and fusion designs in mitigating negative transfer while retaining target-modality discrimination.

\section*{LLM Usage Statement}
\addcontentsline{toc}{section}{LLM Usage Statement}

In this work, we used generative AI tools to assist in editing and refining author-produced manuscript text and implementation code. We have reviewed all AI-assisted work, including verifying the accuracy of the revised text and the correctness of the code. We take responsibility for the final content of this work, including text and code produced with the aid of generative AI.

{\small
  \bibliographystyle{ieeenat_fullname}
  \bibliography{references}
}

\appendix
\pdfbookmark[0]{Appendices}{appendices}
\section{Supplementary Analysis on Logit Space}
\label{appendix:moti}

This section examines logit-space structure and the residual decomposition used in TALD. Section~\ref{appendix:logit_vis} compares PCA projections of same-modality and cross-modal logits and their softmax transformations. Section~\ref{appendix:softmax_property} analyzes the residual formulation, contribution regularization, and confidence dependence of the cross-entropy criteria.

\subsection{Cross-Modal Structural Divergence in the Logit Space}
\label{appendix:logit_vis}

\begin{figure*}[htbp]
	\centering
	\includegraphics[width=\linewidth]{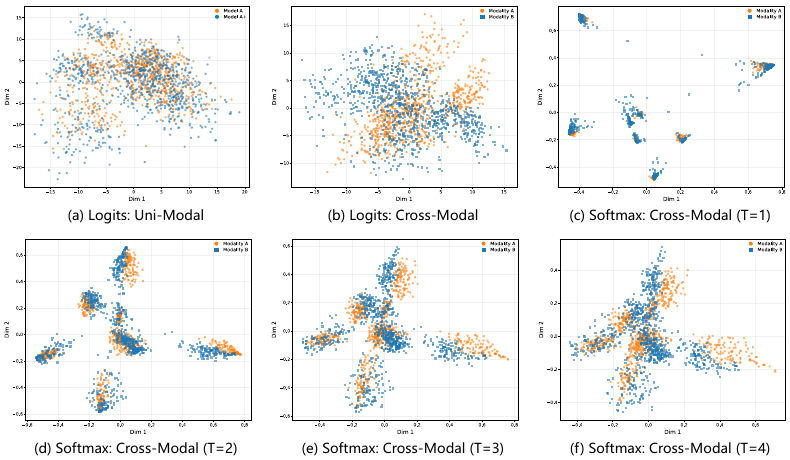}
	\caption{PCA projections of model outputs on RAVDESS.
		(a) Logits from a same-modality model pair.
		(b) Logits from a cross-modal model pair.
		(c)--(f) Softmax probabilities obtained from the cross-modal logits
		at temperatures $T=1$ to $4$.}
	\label{fig:logit_tsne}
\end{figure*}

We use PCA to visualize outputs from same-modality and cross-modal model pairs on RAVDESS, where the logit dimension is $K=8$. Figure~\ref{fig:logit_tsne} compares the patterns visible in the projected logits and their softmax transformations.

\paragraph{Intra- vs.\ cross-modal logit structure.}
In Figure~\ref{fig:logit_tsne}(a)--(b), the same-modality models exhibit similar class-level arrangements in the displayed projections, whereas the cross-modal models show distinct arrangements despite sharing the same class labels. These observations motivate accounting for modality-dependent class-score patterns when constructing the distillation target.

\paragraph{Effect of softmax and temperature.}
Figure~\ref{fig:logit_tsne}(c)--(f) shows that the projected geometry changes after softmax normalization and varies with temperature. For $\mathbf{p}=\mathrm{softmax}(\mathbf{z}/T)$, pairwise logit differences satisfy $z_a-z_b=T\log(p_a/p_b)$. Softmax therefore expresses class-relative scores through probability ratios, while changing the scales and geometry represented by PCA.

Raw logits provide a direct representation of class-relative scores for TALD's additive decomposition. The residual formulation couples the two components to these scores, while C1 and C2 guide their within- and cross-modal classification effects. The learned GEN components are then fused to construct the target for the standard softmax-based KL objective (Section~\ref{sec:distillation}). Section~\ref{appendix:softmax_property} further analyzes the residual formulation and contribution regularization.

\subsection{Logit-Space Structure and the Residual Decomposition}
\label{appendix:softmax_property}

TALD represents each logit vector as an additive combination of modality-general and modality-specific components. The residual parameterization links this decomposition to the original class scores, while the training objective guides the components' classification effects.

\begin{figure}[htbp]
	\centering
	\includegraphics[width=\linewidth]{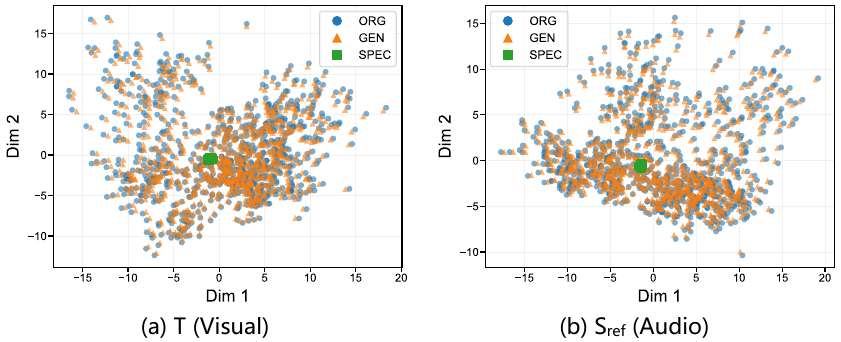}
	\caption{PCA projections of the original logits (ORG),
		modality-general components (GEN), and modality-specific components
		(SPEC) on RAVDESS, shown separately for the teacher (Visual)
		and reference student (Audio). In these projections, SPEC concentrates
		near the origin, while ORG and GEN largely overlap.}
	\label{fig:pca_decomposition}
\end{figure}

\paragraph{Why residual decomposition.}
The residual parameterization enforces $\mathbf{z}^{gen}+\mathbf{z}^{spec}=\mathbf{z}$ exactly. When $\mathbf{z}^{spec}=\mathbf{0}$, the modality-general component recovers the original logits. This coupling provides an explicit reference for learning a correction, whose direction and magnitude are determined during optimization. Figure~\ref{fig:pca_decomposition} shows that the learned SPEC components concentrate near the origin and that ORG and GEN largely overlap in the displayed projections. The module ablation in Table~\ref{tab:module} further supports combining the residual parameterization with sample-adaptive modulation in the evaluated configurations.

\paragraph{Why contribution regularization.}
The contribution measure $\mu_i$ quantifies the relative cross-entropy reduction obtained by adding the modality-specific component to the modality-general logits within its own modality. The regularizer $\mathcal{R}_{\text{ratio}}$ (Eq.~\ref{eq:Rratio}) penalizes contributions outside the sample-specific interval $[\rho_{\min},\rho_{\max}]$ predicted by the Modulator. Since cross-entropy is invariant to uniform shifts of its logit input, this measure focuses on the component's classification effect. The loss ablation in Table~\ref{tab:loss_ablation} shows that adding this regularization improves average accuracy over using the two transfer benefit constraints alone.

\paragraph{Cross-entropy sensitivity at high confidence.}
Under the additive decomposition, both transfer criteria involve the cross-entropy change
$\Delta(\mathbf{v})=\mathrm{CE}(\mathbf{z}^{gen}+\mathbf{v},y)-\mathrm{CE}(\mathbf{z}^{gen},y)$.
For each modality, the C1 hinge has the form
$\mathrm{ReLU}(\Delta(\mathbf{z}^{spec})+\delta)$,
whereas the C2 hinge has the form
$\mathrm{ReLU}(\delta-\Delta(\mathbf{z}^{spec}_{\bar{i}\to i}))$.
The following proposition characterizes how this change depends on the confidence of the base prediction and the size of the perturbation.

\begin{proposition}
	\label{prop:ce_sensitivity}
	Let $\mathbf{z} \in \mathbb{R}^K$ have softmax distribution $\mathbf{p} = \mathrm{softmax}(\mathbf{z})$, true class $c$, and confidence $p_c = 1 - \varepsilon$. For any perturbation $\mathbf{v} \in \mathbb{R}^K$, the cross-entropy gap $\Delta(\mathbf{v}) \triangleq \mathrm{CE}(\mathbf{z} + \mathbf{v}, y) - \mathrm{CE}(\mathbf{z}, y)$ admits the closed form
	\begin{equation}
		\label{eq:ce_gap_form}
		\Delta(\mathbf{v}) = \log\!\Big(p_c + \sum_{j \neq c} p_j\, e^{v_j - v_c}\Big),
	\end{equation}
	and, with $B \triangleq \|\mathbf{v}\|_\infty$, satisfies
	\begin{equation}
		\label{eq:ce_gap_bound}
		\log(1 - \varepsilon) \;\le\; \Delta(\mathbf{v}) \;\le\; \varepsilon\,(e^{2B} - 1).
	\end{equation}
\end{proposition}

\begin{proof}
	Writing cross-entropy in log-sum-exp form, $\mathrm{CE}(\mathbf{z}, y) = -z_c + \log\sum_j e^{z_j}$, we have
	\begin{align*}
		\Delta(\mathbf{v})
		&\overset{\text{(a)}}{=} -v_c + \log\frac{\sum_j e^{z_j + v_j}}{\sum_j e^{z_j}} \\
		&\overset{\text{(b)}}{=} -v_c + \log\sum\nolimits_j p_j\, e^{v_j} \\
		&\overset{\text{(c)}}{=} \log\!\Big(p_c + \sum\nolimits_{j\neq c} p_j\, e^{v_j - v_c}\Big),
	\end{align*}
	where (a) expands both cross-entropies and cancels $-z_c$, (b) substitutes $p_j = e^{z_j}/\sum_k e^{z_k}$, and (c) factors $e^{v_c}$ out of the sum, which establishes \eqref{eq:ce_gap_form}. For the bounds, $\|\mathbf{v}\|_\infty = B$ gives $|v_j - v_c| \le 2B$, hence $e^{-2B} \le e^{v_j - v_c} \le e^{2B}$. Together with $\sum_{j\neq c} p_j = \varepsilon$, this sandwiches the bracketed term,
	\begin{equation*}
		(1-\varepsilon) + \varepsilon e^{-2B}
		\;\le\;
		p_c + \sum_{j\neq c} p_j\, e^{v_j - v_c}
		\;\le\;
		(1-\varepsilon) + \varepsilon e^{2B}.
	\end{equation*}
	Taking logarithms, the upper side gives $\log\!\big(1 + \varepsilon(e^{2B}-1)\big) \le \varepsilon(e^{2B}-1)$ by $\log(1+x) \le x$, and the lower side is at least $\log(1-\varepsilon)$ since $\varepsilon e^{-2B} \ge 0$. This proves Eq.~\eqref{eq:ce_gap_bound}.
\end{proof}

\noindent\textbf{Interpretation.}
For perturbations with uniformly bounded $B$, both bounds approach zero as the base prediction becomes confident in the true class ($\varepsilon\to0$), giving $\Delta(\mathbf{v})\to0$. For C1, the achievable own-modality cross-entropy reduction satisfies
$-\Delta(\mathbf{v})\le-\log(1-\varepsilon)$.
Thus, zero C1 penalty under a fixed positive margin $\delta$ requires
$-\log(1-\varepsilon)\ge\delta$.
For C2, zero hinge penalty with $\delta>0$ requires
$\Delta(\mathbf{v})\ge\delta$, which implies
$B\ge\tfrac{1}{2}\log(1+\delta/\varepsilon)$.

These relations characterize the confidence dependence of the absolute cross-entropy gaps used by C1 and C2, motivating sample-adaptive gap requirements. TALD's predicted margin adjusts the gaps required for zero penalty, while contribution regularization controls relative own-modality effects through the Modulator-predicted interval. The residual parameterization supplies a reference for the decomposition, and the criterion and contribution losses guide the classification effects of its components.

\begin{table*}[htbp]
	\caption{Overview of the five multimodal datasets utilized in our
		experiments, including modality composition, data scale, number of classes,
		and target task for each dataset.}
	\label{tab:appendix_dataset}
	\centering
	{%
		\fontsize{10}{12}\selectfont
		\rmfamily
		\renewcommand{\arraystretch}{1.0}
		\setlength{\tabcolsep}{4mm}

		\begin{tabular}{@{}llccl@{}}
			\toprule
			\textbf{Dataset}
			& \textbf{Modality}
			& \textbf{Scale}
			& \textbf{Classes}
			& \textbf{Task} \\
			\midrule

			AV-MNIST
			& Image, Audio
			& 70k
			& 10
			& Handwritten digit recognition \\

			RAVDESS (Speech)
			& Video, Audio
			& 1,440
			& 8
			& Emotion recognition \\

			VGGSound-50k
			& Video, Audio
			& 49k
			& 141
			& Scene classification \\

			CrisisMMD-V2
			& Image, Text
			& 16k
			& 8
			& Humanitarian classification \\

			Fakeddit
			& Image, Text
			& 50k
			& 6
			& Fine-grained fake news detection \\
			\bottomrule
		\end{tabular}
	}

\end{table*}

\begin{table*}[htbp]
	\caption{Unimodal model configurations.
		I, V, A, and T denote image, video, audio, and text modalities,
		respectively.}
	\label{tab:appendix_model_cfg}
	\centering
	{%
		\fontsize{9}{11}\selectfont
		\rmfamily
		\renewcommand{\arraystretch}{1.15}
		\setlength{\tabcolsep}{3mm}

		\begin{tabular}{@{}lll@{}}
			\toprule
			\textbf{Dataset}
			& \textbf{Unimodal Model 1}
			& \textbf{Unimodal Model 2} \\
			\midrule
			AV-MNIST
			& I: LeNet5
			& A: ThreeLayerCNN-2D \\

			RAVDESS
			& V: ThreeLayerCNN-3D + MLP
			& A: ThreeLayerCNN-1D + MLP \\

			VGGSound-50k
			& V: ThreeLayerCNN-3D + MLP
			& A: ThreeLayerCNN-1D + MLP \\

			CrisisMMD-V2 / Fakeddit
			& I: MLP
			& T: MLP \\
			\bottomrule
		\end{tabular}
	}
\end{table*}

\section{Supplementary Dataset and Implementation Details}
\label{appendix:moreI&E}

This section provides additional details on the experimental setup. Section~\ref{appendix:dataset} describes the five multimodal datasets, Section~\ref{appendix:model} specifies the model configurations used for each dataset, and Section~\ref{appendix:training_cfg} lists the training configurations.

\subsection{Dataset Descriptions}
\label{appendix:dataset}
For the multimodal datasets used in the main text, we provide a summary of their key characteristics in Table~\ref{tab:appendix_dataset}, followed by detailed descriptions in the subsequent paragraphs.

\textbf{AV-MNIST:}
The AV-MNIST dataset is a multimodal benchmark for digit recognition, comprising 70,000 paired samples across 10 classes. The visual modality consists of $28 \times 28$ MNIST images with 75\% of their energy removed via principal component analysis (PCA) to simulate low-quality input. The auditory modality is derived from spoken digit audio samples of the Tidigits database, augmented by overlaying noise segments from the ESC-50 dataset~\cite{esc}, and represented as $112 \times 112$ spectrograms.

\textbf{RAVDESS:}
The RAVDESS dataset is a multimodal benchmark for emotion recognition, comprising 1,440 audiovisual samples of emotional utterances across 8 classes. Both video and audio modalities are preprocessed to a uniform duration of 3.6 seconds. Visual inputs are derived from 15 uniformly sampled frames, with facial regions extracted using MTCNN~\cite{mtcnn}. The audio modality is represented by 15-dimensional Mel-frequency cepstral coefficients (MFCCs) extracted at a sampling rate of 22,050 Hz.

\textbf{VGGSound-50k:}
The VGGSound-50k dataset is a multimodal benchmark comprising 48,755 YouTube video clips across 141 fine-grained audio-visual scene categories, derived from VGGSound-AVEL50k~\cite{cpsp} through quality filtering. Visual features are extracted using VGG19~\cite{vgg19} at 16 fps (512D) and audio features via VGGish~\cite{vggish} with standard preprocessing (128D), maintaining temporal alignment between modalities.

\textbf{CrisisMMD-V2:}
The CrisisMMD-V2 dataset is a multimodal benchmark for humanitarian crisis classification, containing 16,058 annotated Twitter posts, each comprising a paired image and text, spanning 8 humanitarian categories. For feature extraction, we utilize a pretrained ResNet-50~\cite{resnet} (with the classification head removed) for images and BERTweet-base~\cite{bertweet} for textual representations.

\textbf{Fakeddit:}
The Fakeddit dataset is a large-scale multimodal benchmark for fine-grained fake news detection, sourced from Reddit and providing 2-way, 3-way, and 6-way classification labels through distant supervision. We adopt the 6-way label scheme and downsample the dataset to 49,982 samples while preserving the original class proportions. For feature extraction, we utilize a pretrained ResNet-50 (with the classification head removed) for images, yielding 2,048-dimensional visual features, and BERT-base-uncased~\cite{bert} for text, using the \texttt{[CLS]} token output as 768-dimensional textual representations.

\subsection{Model Configurations Across Datasets}
\label{appendix:model}

The unimodal architectures are summarized in Table~\ref{tab:appendix_model_cfg}. We follow the unimodal model configurations of~\cite{mst-distill} for the shared benchmarks and extend the implementation to six-way Fakeddit using separate image and text MLP classifiers. All compared methods use the same corresponding teacher and student architectures within each dataset.

Our evaluation focuses on transfer from a unimodal source teacher. TALD additionally uses an independently trained target-modality reference student. MST-Distill employs a multimodal teacher together with an auxiliary unimodal teacher and a different training protocol; its original configuration is therefore outside the scope of this comparison. Optimization settings are provided in Section~\ref{appendix:training_cfg}.

\subsection{Training Configuration}
\label{appendix:training_cfg}

The common unimodal training protocol uses Adam without weight decay and fixed learning rates. AV-MNIST, VGGSound-50k, and Fakeddit use a learning rate of $10^{-4}$ and a batch size of 512; RAVDESS uses $10^{-4}$ and 32; and CrisisMMD-V2 uses $5\times10^{-3}$ and 512. These dataset-specific optimization settings are shared across compared methods. For Fakeddit, both the image and text MLP classifiers are trained for 100 epochs with random seeds $\{0,1,2,3,4\}$.

For TALD, the source teacher and target-modality reference student are trained independently with supervised classification losses and frozen before Stage~1. For each dataset and seed, the two unimodal models are trained once and reused across both transfer directions: each serves as the source teacher in one direction and the target-modality reference student in the other. Stage~1 optimizes the Disentangler and Modulator for up to 30 epochs, with early stopping based on validation loss. Stage~2 freezes these modules and trains the final student for 100 epochs. The distillation temperature is $T=4$, and the distillation weight $\alpha$ starts at 0.5 and is halved every 30 epochs.

TALD uses $\beta=5$ and $[\delta_{\min},\delta_{\max}]=[-0.1,0.1]$. The ratio bounds are mapped to $[\rho_{\min}^{-},\rho_{\min}^{+}]=[0.05,0.2]$ and $[\rho_{\max}^{-},\rho_{\max}^{+}]=[0.3,0.8]$. The loss weights are $\lambda_{\mathrm{hyp}}=\lambda_{\mathrm{ratio}}=\lambda_{\mathrm{aux}}=1$, $\lambda_{\mathrm{con}}=0.1$, and $\lambda_{\mathrm{var}}=0.01$, with contrastive temperature $\tau_{\mathrm{con}}=0.1$. These TALD-specific defaults are shared across all five datasets. All experiments are conducted on an Intel Xeon Gold 6248R CPU with an NVIDIA A100 GPU.

\section{Supplementary TALD Experiments}

This section provides additional experiments and analyses of TALD.
We first report the complete distillation target ablation
(Section~\ref{appendix:target_ablation}).
We then examine frozen-checkpoint CMP diagnostics
(Section~\ref{appendix:cmp}), evaluate the auxiliary regularization
(Section~\ref{appendix:aux_reg}), and analyze the signed margin range
(Section~\ref{appendix:margin_range}).
We also visualize the disentangled logit components
(Section~\ref{appendix:disentangle_vis}) and report complete
stage-wise and end-to-end costs for all
compared methods across the five datasets
(Section~\ref{appendix:efficiency}).
Section~\ref{appendix:full_dynamics} presents the complete
training dynamics. Section~\ref{appendix:positive_transfer_frequency}
jointly analyzes mean accuracy gains and positive-transfer frequencies.

\subsection{Complete Ablation of Distillation Targets}
\label{appendix:target_ablation}

Table~\ref{tab:target_ablation_full} reports the complete set of
Stage~2 target configurations underlying
Table~\ref{tab:target_ablation}. Results are averaged over five
runs. Each dataset entry averages the two transfer directions,
and Avg denotes the mean across the two datasets.

\paragraph{Target construction.}
ORG, GEN, and SPEC denote original, modality-general, and
modality-specific logits, respectively. SPEC targets are matched
to their GEN counterparts in root mean square (RMS) magnitude
using fixed scales estimated from centered training logits.
Shuffled-S and Shuffled-C retain the GEN
logits and permute the learned fusion weights across samples
and within classes, respectively.

\paragraph{Results.}
Adaptive GEN fusion achieves the highest accuracy on both
datasets and outperforms GEN distillation from either source
alone. GEN yields a higher cross-dataset average than SPEC in
each matched source/fusion setting, although SPEC is slightly
better on Fakeddit for teacher-only distillation and mean fusion.
The Shuffled-S and Shuffled-C GEN targets obtain
averages of 0.6400 and 0.6410, respectively, compared with
0.6394 for equal weighting and 0.6466 for adaptive fusion.
The lower accuracy under both shuffling controls supports
the utility of associating each learned weight with its
corresponding sample, including among samples of the same class.

\begin{table}[t]
	\caption{Complete distillation target ablation.
		ORG, GEN, and SPEC denote original, modality-general,
		and modality-specific logits. Both denotes $T$ and
		$S_{\mathrm{ref}}$. Mean and Adaptive use equal weights
		and the learned weight $w$, respectively.
		Shuffled-S permutes $w$ across samples;
		Shuffled-C restricts the permutation to each class.
		The best and second-best results per column are
		\textbf{bold}, with the best additionally
		\textbf{\underline{underlined}}.}
	\label{tab:target_ablation_full}
	\centering
	{%
		\fontsize{9}{11}\selectfont
		\rmfamily
		\renewcommand{\arraystretch}{0.85}
		\setlength{\tabcolsep}{0.65mm}
		\begin{tabular}{@{}ccccccc@{}}
			\toprule
			\textbf{\#} & \textbf{Logits} & \textbf{Source} & \textbf{Fusion} & \textbf{RAVDESS} & \textbf{Fakeddit} & \textbf{Avg} \\
			\midrule
			(a) & GEN & $S_{\mathrm{ref}}$ & \textemdash
			& 0.7333 & 0.5436 & 0.6385 \\
			(b) & GEN & $T$ & \textemdash
			& 0.7021 & 0.5442 & 0.6232 \\
			(c) & SPEC & $S_{\mathrm{ref}}$ & \textemdash
			& 0.7222 & 0.5428 & 0.6325 \\
			(d) & SPEC & $T$ & \textemdash
			& 0.6962 & 0.5446 & 0.6204 \\
			\midrule
			(e) & GEN & Both & Mean
			& 0.7358 & 0.5430 & 0.6394 \\
			(f) & SPEC & Both & Mean
			& 0.7167 & \textbf{0.5447} & 0.6307 \\
			\midrule
			(g) & ORG & Both & Adaptive
			& \textbf{0.7410} & 0.5443 & \textbf{0.6427} \\
			(h) & SPEC & Both & Adaptive
			& 0.7247 & 0.5428 & 0.6338 \\
			\midrule
			(i) & GEN & Both & Shuffled-S
			& 0.7368 & 0.5431 & 0.6400 \\
			(j) & GEN & Both & Shuffled-C
			& 0.7375 & 0.5445 & 0.6410 \\
			\rowcolor{gray!10}
			\textbf{(k)} & GEN & Both & Adaptive
			& \textbf{\underline{0.7448}} & \textbf{\underline{0.5483}} & \textbf{\underline{0.6466}} \\
			\bottomrule
		\end{tabular}
	}
\end{table}

\subsection{Frozen-Checkpoint Diagnostics of CMP}
\label{appendix:cmp}

The downstream ablation in Table~\ref{tab:cmp_ablation} evaluates
CMP parameterization through student accuracy. We complement it
with frozen-checkpoint interventions that examine how the
SPEC-dependent term, class-conditioned bias, sample pairing, and
learned margins affect the C2 proxy. Together, these analyses assess
CMP's downstream utility and the behavior of the constraint used
to guide decomposition.

\paragraph{Frozen-checkpoint protocol.}
We freeze 20 trained checkpoints of the class-conditioned affine model: two datasets, two checkpoint tasks per dataset, and five seeds. Each checkpoint contains two directional CMPs, yielding 40 evaluated mappings. We use each checkpoint's complete original test split: 288 samples on RAVDESS and 9,997 on Fakeddit. We retain the original seed-specific data splits. No model is retrained or selected using these diagnostics. Conditions share recipient GEN logits, labels, margins, and parameters. Test labels select class-conditioned CMP parameters only for this diagnostic, not for student prediction 	or parameter fitting. Statistics are averaged over samples, then over the same 20 saved within-class donor permutations where applicable, and finally equally over five seeds. Each saved permutation is a within-class derangement of the complete test split, and all samples are eligible. We report the sample standard deviation of the five seed-level values; permutations and the two mappings within a checkpoint are not independent model runs. The cross-seed spread therefore includes variation associated with the original splits and should not be attributed solely to optimization randomness.

\paragraph{Interventions and diagnostic.}
For source SPEC $\mathbf{s}_i$, recipient GEN $\mathbf{g}_i$, and
label $y_i$, define
\[
G_i(\mathbf{u})=
\mathrm{CE}(\mathbf{g}_i+\mathbf{u},y_i)
-\mathrm{CE}(\mathbf{g}_i,y_i).
\]
Cross-entropy is evaluated at temperature one, in nats.
Positive $G_i$ denotes an increase in recipient cross-entropy after
the frozen CMP contribution. We compare
\[
\begin{aligned}
	\mathbf{u}_i^{\mathrm{O}}
	&=\mathbf{s}_i\mathbf{W}^{(y_i)}+\mathbf{b}^{(y_i)},\\
	\mathbf{u}_i^{\mathrm{Z}}&=\mathbf{b}^{(y_i)},\\
	\mathbf{u}_i^{\mathrm{L}}&=\mathbf{s}_i\mathbf{W}^{(y_i)},\\
	\mathbf{u}_i^{\pi}
	&=\mathbf{s}_{\pi(i)}\mathbf{W}^{(y_i)}+\mathbf{b}^{(y_i)}.
\end{aligned}
\]
Here, $\mathrm{Z}$ zeros only the CMP's SPEC input, $\mathrm{L}$
removes only its additive bias, and $\pi$ reassigns source SPEC within
the same class while retaining the recipient quantities.
Write $G_i^{a}=G_i(\mathbf{u}_i^{a})$ and
$\Delta G_i=G_i^{\pi}-G_i^{\mathrm{O}}$.
This SPEC permutation differs from the fusion-weight permutations
in Table~\ref{tab:target_ablation_full}. The frozen SPEC-zero
intervention also differs from the independently trained
class-conditioned bias-only variant (d) in Table~\ref{tab:cmp_ablation}.

\paragraph{SPEC-dependent and bias effects.}
Table~\ref{tab:cmp_frozen} reports the saved original GPU statistics.
Across all eight configurations, removing either the SPEC input or
the bias reduces the original mean gap. Under SPEC-zero, the mean
gap remains positive and no negative $G$ is observed in either
numerical path, showing that the learned bias alone can induce the
desired cross-entropy increase. Retaining the SPEC-dependent term
further increases the mean gap in every configuration. Both terms
therefore contribute to the fitted cross-modal response, although
the positive SPEC-zero gaps also show why C2 satisfaction alone
cannot establish input-dependent exclusivity. Since cross-entropy
is nonlinear, these gap reductions are not additive attributions.

\paragraph{Within-class pairing effects.}
Within-class shuffling preserves the source SPEC distribution and
class-conditioned parameters while changing the source--recipient
correspondence. The mean absolute gap changes are
$(1.807\text{--}1.929)\times10^{-3}$ nats on RAVDESS and
$(7.833\text{--}10.025)\times10^{-3}$ nats on Fakeddit, whereas the
signed mean changes are much smaller. Positive and negative
responses thus largely cancel in the group mean. Because the bias
is unchanged by these permutations, the observed variation
demonstrates sensitivity to SPEC--recipient pairing beyond the
class-conditioned offset. This complements the component-removal
diagnostics by testing sample correspondence within each class.

\begin{table*}[t]
	\centering
	\caption{Frozen CMP interventions, using saved original GPU values.
		All five numerical columns are in $10^{-3}$ nats and report mean
		$\pm$ sample standard deviation across five seeds after within-seed
		aggregation. Under Direction, Distillation denotes the teacher-to-student
		direction used to train the checkpoint, and CMP denotes the source-SPEC
		to recipient-GEN mapping evaluated within it. Each checkpoint contains
		both CMP directions. V/A and I/T denote visual/audio and image/text,
		respectively. SPEC=0 changes only the CMP input; Bias=0 changes only
		the CMP additive bias. No model is retrained.}
	\label{tab:cmp_frozen}
	\begingroup
	\fontsize{9}{11}\selectfont
	\rmfamily
	\setlength{\tabcolsep}{4pt}
	\renewcommand{\arraystretch}{1.15}
	\edef\CompactHeaderShift{%
		\the\dimexpr0pt-(0.4ex+\cmidrulewidth)/2\relax}
	\edef\CMPDatasetShift{%
		\the\dimexpr0pt-(\aboverulesep+\belowrulesep+\cmidrulewidth)/2\relax}
	\begin{tabular*}{\textwidth}{@{\extracolsep{\fill}}cccccccc@{}}
		\toprule
		\multirow[c]{2}{*}[\CompactHeaderShift]{\textbf{Dataset}}
		& \multicolumn{2}{c}{\textbf{Direction}}
		& \multicolumn{3}{c}{\textbf{Mean} $G$}
		& \multicolumn{2}{c}{\textbf{Within-class shuffle}} \\
		\noalign{\vskip\dimexpr0.2ex-\aboverulesep\relax}
		\cmidrule(lr){2-3}\cmidrule(lr){4-6}\cmidrule(lr){7-8}
		\noalign{\vskip\dimexpr0.2ex-\belowrulesep\relax}
		& Distillation & CMP
		& Original & SPEC=0 & Bias=0
		& $\overline{\Delta G}$ & $\overline{|\Delta G|}$ \\
		\midrule
		\multirow[c]{4}{*}[\CMPDatasetShift]{\rotatebox[origin=c]{90}{\textbf{RAVDESS}}} & \multirow[c]{2}{*}{$\mathrm{V}\!\to\!\mathrm{A}$} & $\mathrm{V}\!\to\!\mathrm{A}$
		& $4.106\pm3.220$ & $2.688\pm2.022$ & $1.412\pm1.206$ & $0.142\pm0.103$ & $1.807\pm0.241$ \\
		&  & $\mathrm{A}\!\to\!\mathrm{V}$
		& $4.548\pm3.046$ & $2.492\pm1.685$ & $2.047\pm1.421$ & $0.074\pm0.147$ & $1.929\pm0.309$ \\
		\cmidrule(lr){2-8}
		& \multirow[c]{2}{*}{$\mathrm{A}\!\to\!\mathrm{V}$} & $\mathrm{A}\!\to\!\mathrm{V}$
		& $4.298\pm3.186$ & $2.639\pm1.984$ & $1.651\pm1.295$ & $0.062\pm0.211$ & $1.856\pm0.324$ \\
		&  & $\mathrm{V}\!\to\!\mathrm{A}$
		& $4.307\pm2.970$ & $2.555\pm1.846$ & $1.747\pm1.119$ & $0.077\pm0.169$ & $1.883\pm0.192$ \\
		\midrule
		\multirow[c]{4}{*}[\CMPDatasetShift]{\rotatebox[origin=c]{90}{\textbf{Fakeddit}}} & \multirow[c]{2}{*}{$\mathrm{I}\!\to\!\mathrm{T}$} & $\mathrm{I}\!\to\!\mathrm{T}$
		& $6.174\pm3.306$ & $4.160\pm0.734$ & $2.011\pm2.630$ & $-0.104\pm0.137$ & $10.002\pm1.258$ \\
		&  & $\mathrm{T}\!\to\!\mathrm{I}$
		& $4.807\pm3.256$ & $3.469\pm0.698$ & $1.335\pm2.707$ & $0.021\pm0.091$ & $7.923\pm0.883$ \\
		\cmidrule(lr){2-8}
		& \multirow[c]{2}{*}{$\mathrm{T}\!\to\!\mathrm{I}$} & $\mathrm{T}\!\to\!\mathrm{I}$
		& $4.809\pm2.643$ & $3.643\pm0.713$ & $1.163\pm2.014$ & $-0.051\pm0.162$ & $7.833\pm1.804$ \\
		&  & $\mathrm{I}\!\to\!\mathrm{T}$
		& $7.806\pm3.245$ & $4.265\pm0.846$ & $3.535\pm2.542$ & $-0.007\pm0.145$ & $10.025\pm0.678$ \\
		\bottomrule
	\end{tabular*}
	\endgroup
\end{table*}

\paragraph{Threshold-sensitive pairing diagnostics.}
Assign $\ell_\tau(G)=+1$ for $G>\tau$, $-1$ for $G<-\tau$,
and $0$ otherwise. For an original--shuffled pair, let
$a_i=\ell_\tau(G_i^{\mathrm{O}})$ and
$b_i=\ell_\tau(G_i^{\pi})$. Define empirical fractions
\[
\begin{aligned}
	r_\tau&=\Pr(a_i b_i=-1),\\
	q_\tau&=\Pr(\{a_i,b_i\}=\{0,1\}),\\
	c_\tau&=\Pr(|a_i b_i|=1),\\
	d_\tau&=\Pr(\mathbf{1}_{a_i=1}\ne\mathbf{1}_{b_i=1}).
\end{aligned}
\]
All denominators include every paired sample, including near-zero
pairs. Robust sign flips $r_\tau$ and positive--near-zero transitions
$q_\tau$ partition the positive-indicator change rate:
$d_\tau=r_\tau+q_\tau$. Negative--near-zero transitions are excluded
from $d_\tau$. This identity holds within each permutation and seed
and for their means; it does not imply that standard deviations add.
Coverage $c_\tau$ reports the fraction with both
values outside the near-zero band; $r_\tau$ is not divided by this
coverage.
Since $q_\tau$ counts only positive--near-zero transitions,
$q_\tau\leq 1-c_\tau$, where $1-c_\tau$ is the fraction of pairs
with at least one near-zero value. The remaining such pairs are
negative--near-zero or near-zero--near-zero pairs.
Table~\ref{tab:cmp_robustness} reports three thresholds.
At $\tau=10^{-4}$, the seed-averaged robust sign-flip rates span
8.46\%--8.98\% on RAVDESS and 21.77\%--28.54\% on Fakeddit.
Pairing sensitivity therefore persists when both gaps lie outside
the near-zero band, rather than being confined to near-zero
transitions. These rates characterize the learned response;
larger values do not imply better transfer.

\begin{table*}[t]
	\centering
	\caption{Pairing diagnostics from full float64 recomputation of the
		cached logits and stored CMP parameters, using the same samples and
		donor permutations. Values are percentages, reported as mean $\pm$
		sample standard deviation over five seeds. Robust sign-flip rates
		$r_\tau$ are given at three tolerances. The diagnostic columns at
		$\tau=10^{-4}$ report positive--near-zero transitions ($q_\tau$)
		and the fraction of pairs with both values outside the near-zero
		band ($c_\tau$). Every rate uses all paired samples as its denominator.}
	\label{tab:cmp_robustness}
	\begingroup
	\fontsize{9}{11}\selectfont
	\rmfamily
	\setlength{\tabcolsep}{4pt}
	\renewcommand{\arraystretch}{1.15}
	\edef\CompactHeaderShift{%
		\the\dimexpr0pt-(0.4ex+\cmidrulewidth)/2\relax}
	\edef\CMPDatasetShift{%
		\the\dimexpr0pt-(\aboverulesep+\belowrulesep+\cmidrulewidth)/2\relax}
	\begin{tabular*}{\textwidth}{@{\extracolsep{\fill}}cccccccc@{}}
		\toprule
		\multirow[c]{2}{*}[\CompactHeaderShift]{\textbf{Dataset}}
		& \multicolumn{2}{c}{\textbf{Direction}}
		& \multicolumn{3}{c}{\textbf{Robust sign flip} $r_\tau$}
		& \multicolumn{2}{c}{\textbf{Pair diagnostics} ($\tau=10^{-4}$)} \\
		\noalign{\vskip\dimexpr0.2ex-\aboverulesep\relax}
		\cmidrule(lr){2-3}\cmidrule(lr){4-6}\cmidrule(lr){7-8}
		\noalign{\vskip\dimexpr0.2ex-\belowrulesep\relax}
		& Distillation & CMP
		& $\tau=10^{-6}$ & $\tau=10^{-5}$ & $\tau=10^{-4}$
		& Pos.--near-zero $q_\tau$ & Both non-near-zero $c_\tau$ \\
		\midrule
		\multirow[c]{4}{*}[\CMPDatasetShift]{\rotatebox[origin=c]{90}{\textbf{RAVDESS}}} & \multirow[c]{2}{*}{$\mathrm{V}\!\to\!\mathrm{A}$} & $\mathrm{V}\!\to\!\mathrm{A}$
		& $12.33\pm9.51$ & $11.18\pm8.54$ & $8.46\pm6.57$ & $3.70\pm1.50$ & $82.24\pm8.92$ \\
		&  & $\mathrm{A}\!\to\!\mathrm{V}$
		& $12.11\pm12.83$ & $11.39\pm11.97$ & $8.78\pm8.81$ & $4.16\pm2.46$ & $85.68\pm8.67$ \\
		\cmidrule(lr){2-8}
		& \multirow[c]{2}{*}{$\mathrm{A}\!\to\!\mathrm{V}$} & $\mathrm{A}\!\to\!\mathrm{V}$
		& $13.55\pm9.21$ & $12.39\pm8.31$ & $8.98\pm6.54$ & $4.22\pm1.78$ & $84.32\pm7.52$ \\
		&  & $\mathrm{V}\!\to\!\mathrm{A}$
		& $11.31\pm9.16$ & $10.55\pm8.46$ & $8.49\pm6.99$ & $4.00\pm1.57$ & $84.49\pm6.58$ \\
		\midrule
		\multirow[c]{4}{*}[\CMPDatasetShift]{\rotatebox[origin=c]{90}{\textbf{Fakeddit}}} & \multirow[c]{2}{*}{$\mathrm{I}\!\to\!\mathrm{T}$} & $\mathrm{I}\!\to\!\mathrm{T}$
		& $29.27\pm6.20$ & $29.20\pm6.19$ & $28.54\pm6.12$ & $1.03\pm0.06$ & $98.43\pm0.25$ \\
		&  & $\mathrm{T}\!\to\!\mathrm{I}$
		& $30.52\pm9.74$ & $30.01\pm9.63$ & $27.88\pm9.06$ & $2.73\pm0.43$ & $93.11\pm0.68$ \\
		\cmidrule(lr){2-8}
		& \multirow[c]{2}{*}{$\mathrm{T}\!\to\!\mathrm{I}$} & $\mathrm{T}\!\to\!\mathrm{I}$
		& $29.83\pm10.80$ & $29.47\pm10.66$ & $27.42\pm9.82$ & $3.24\pm0.64$ & $93.52\pm1.88$ \\
		&  & $\mathrm{I}\!\to\!\mathrm{T}$
		& $22.27\pm8.35$ & $22.22\pm8.34$ & $21.77\pm8.20$ & $0.70\pm0.25$ & $98.97\pm0.37$ \\
		\bottomrule
	\end{tabular*}
	\endgroup
\end{table*}

\paragraph{Numerical verification.}
For Table~\ref{tab:cmp_robustness}, cached logits and CMP parameters
are cast to float64 before recomputing the affine projection,
logit addition, and cross-entropy. Relative to the saved GPU
diagnostics, the maximum absolute changes in per-seed mean $G$
across all four conditions and in shuffle mean absolute
$\Delta G$ are $2.053936\times10^{-8}$ and
$6.327671\times10^{-9}$ nats. The largest sample-level discrepancies
in $G$ and paired $\Delta G$ are $2.104056\times10^{-6}$ and
$1.818920\times10^{-6}$ nats. The maximum net change in seed-averaged
$d_{10^{-6}}$ is 0.013889 percentage points, although individual
threshold labels can disagree. These are observed differences,
not numerical error bounds. The original float32 CMP outputs and
perturbed logits were not retained, and recomputation also changes
the CPU/GPU backend and PyTorch version, so cross-entropy precision
alone cannot be isolated.

\paragraph{Margins and interpretation.}
All cached margins are negative: their observed ranges are
$[-0.090550,-0.008204]$ on RAVDESS and
$[-0.099912,-0.097109]$ on Fakeddit. The original mean directional
hinge $[\delta_i-G_i^{\mathrm{O}}]_+$ is zero for all 40 evaluated
mappings. Nevertheless, the five-seed configuration means of
$[-G_i^{\mathrm{O}}]_+$ range from
$2.22\times10^{-4}$ to $2.87\times10^{-4}$ nats on RAVDESS and
from $1.56\times10^{-3}$ to $2.28\times10^{-3}$ nats on Fakeddit.
Thus, satisfying the relaxed hinge does not establish universal
zero-margin exclusivity. Algebraically, $G=0$ also satisfies this
hinge when $\delta<0$; this is not an additional measured
intervention. These frozen test observations neither imply that
C2 supplied no training gradients nor replace the separate
epoch-wise observations in Appendix~\ref{appendix:full_dynamics}.

\paragraph{Implications for CMP design.}
The frozen diagnostics identify class-dependent offsets and
SPEC-dependent, pairing-sensitive responses within the learned
affine mapping. Their downstream utility is supported by the
training ablations: the full class-conditioned affine CMP reaches
Avg 0.6466, compared with 0.6434 for class-independent affine,
0.6450 for class-conditioned linear, and 0.6430 for bias-only
(Table~\ref{tab:cmp_ablation}). Adding C2 to C1 with both
regularizers retained further improves accuracy by 0.38 and
0.05 percentage points on RAVDESS and Fakeddit, respectively
(Table~\ref{tab:loss_ablation}). CMP guides SPEC through the
Stage-1 cross-modal constraint; residual coupling links these
changes to the GEN components used in Stage~2. Together, the
results support class- and sample-sensitive guidance for
task-dependent target construction, consistent with
Section~\ref{sec:criteria_status}. They do not establish strict
semantic separation or isolate pairing sensitivity as the cause
of the accuracy gains.

\subsection{Auxiliary Regularization: Formulation and Ablation}
\label{appendix:aux_reg}

\begin{table}[htbp]
	\caption{Ablation of auxiliary regularization
		$\mathcal{R}_{\mathrm{aux}}$ on VGGSound-50k and Fakeddit, averaged over five independent runs. $\mathcal{L}_{\mathrm{con}}$ denotes the supervised contrastive loss, and $\mathcal{L}_{\mathrm{var}}$ denotes the margin variance regularization.}
	\label{tab:aux_ablation}
	\centering
	{%
		\fontsize{9}{11}\selectfont
		\rmfamily
		\renewcommand{\arraystretch}{1.0}
		\setlength{\tabcolsep}{0.65mm}

		\edef\CompactHeaderShift{%
			\the\dimexpr0pt-(0.4ex+\cmidrulewidth)/2\relax}
		\begin{tabular}{lccccccc}
			\toprule
			\multirow[c]{2}{*}[\CompactHeaderShift]{\textbf{\#}}
			& \multicolumn{2}{c}{$\mathcal{R}_{\mathrm{aux}}$}
			& \multicolumn{2}{c}{\textbf{VGGSound-50k}}
			& \multicolumn{2}{c}{\textbf{Fakeddit}}
			& \multirow[c]{2}{*}[\CompactHeaderShift]{\textbf{Avg}} \\
			\noalign{\vskip\dimexpr0.2ex-\aboverulesep\relax}
			\cmidrule(lr){2-3}
			\cmidrule(lr){4-5}
			\cmidrule(lr){6-7}
			\noalign{\vskip\dimexpr0.2ex-\belowrulesep\relax}
			& $\mathcal{L}_{\mathrm{con}}$
			& $\mathcal{L}_{\mathrm{var}}$
			& M1(V)
			& M2(A)
			& M1(I)
			& M2(T)
			& \\
			\midrule

			(a)
			& \ding{55}
			& \ding{55}
			& 0.4464
			& \textbf{0.5946}
			& \textbf{\underline{0.5773}}
			& 0.5158
			& 0.5335 \\

			(b)
			& \ding{51}
			& \ding{55}
			& \textbf{0.4493}
			& \textbf{\underline{0.5947}}
			& 0.5759
			& \textbf{0.5184}
			& \textbf{0.5346} \\

			(c)
			& \ding{55}
			& \ding{51}
			& \textbf{\underline{0.4496}}
			& 0.5940
			& 0.5763
			& 0.5145
			& 0.5336 \\

			\rowcolor{gray!10}
			\textbf{(d) Ours}
			& \ding{51}
			& \ding{51}
			& 0.4484
			& 0.5939
			& \textbf{0.5772}
			& \textbf{\underline{0.5194}}
			& \textbf{\underline{0.5347}} \\
			\bottomrule
		\end{tabular}
	}

\end{table}

The two auxiliary terms introduced in Section~\ref{sec:objective} are defined as follows. The supervised contrastive loss~\cite{scl} on the hidden representation $\mathbf{h}$ is
\begin{equation}
	\resizebox{0.85\columnwidth}{!}{$
		\mathcal{L}_{\text{con}} = \frac{1}{|\mathcal{V}|} \sum_{i \in \mathcal{V}} \frac{-1}{|\mathcal{P}(i)|} \sum_{p \in \mathcal{P}(i)} \log \frac{\exp(\cos(\mathbf{h}_i, \mathbf{h}_p) / \tau_{\text{con}})}{\sum_{j \neq i} \exp(\cos(\mathbf{h}_i, \mathbf{h}_j) / \tau_{\text{con}})},
		$}
	\label{eq:Lcon}
\end{equation}
where $\mathcal{P}(i)$ denotes same-class positives, $\mathcal{V}$ is the set of valid anchors, and $\tau_{\text{con}} = 0.1$ is the contrastive temperature.

The variance regularizer penalizes the within-batch variance of the predicted margins when it falls outside a prescribed interval:
\begin{equation}
	\resizebox{0.85\columnwidth}{!}{$
		\mathcal{L}_{\mathrm{var}}
		=
		\mathrm{ReLU}\!\left(
		0.001-\mathrm{Var}_{\mathcal{B}}(\boldsymbol{\delta})
		\right)
		+
		\mathrm{ReLU}\!\left(
		\mathrm{Var}_{\mathcal{B}}(\boldsymbol{\delta})-0.01
		\right),
		$}
	\label{eq:Lvar}
\end{equation}
where $\mathrm{Var}_{\mathcal{B}}$ denotes the variance within a mini-batch.

Table~\ref{tab:aux_ablation} evaluates $\mathcal{R}_{\mathrm{aux}}$
on VGGSound-50k and Fakeddit. The full combination achieves the
highest reported Avg (0.5347), compared with 0.5335 without
auxiliary regularization and 0.5346 with
$\mathcal{L}_{\mathrm{con}}$ alone.
The effects vary across transfer directions:
$\mathcal{L}_{\mathrm{con}}$ alone improves M2 on Fakeddit from
0.5158 to 0.5184, whereas $\mathcal{L}_{\mathrm{var}}$ alone
improves M1 on VGGSound-50k from 0.4464 to 0.4496.
The combined configuration obtains the highest Fakeddit M2
accuracy (0.5194), although it does not improve every direction.
The weights $\lambda_{\mathrm{con}}=0.1$ and
$\lambda_{\mathrm{var}}=0.01$ are shared across all five datasets
without per-dataset tuning. These results support a modest
average benefit of the auxiliary regularization in the evaluated
settings.

\begin{table}[t]
	\caption{Ablation on the margin range
		$[\delta_{\mathrm{min}},\delta_{\mathrm{max}}]$ on RAVDESS,
		averaged over five independent runs.
		$\delta_{--}$ and $\delta_{-}$ restrict the margin to
		non-positive ranges, whereas $\delta_{+}$ and $\delta_{++}$
		restrict it to non-negative ranges.
		The proposed setting $\delta$ spans both negative and
		positive values.}
	\label{tab:margin_range}
	\centering
	{%
		\fontsize{9}{11}\selectfont
		\rmfamily
		\renewcommand{\arraystretch}{1.0}
		\setlength{\tabcolsep}{1.5mm}

		\begin{tabular}{llccc}
			\toprule
			\textbf{Setting}
			& \textbf{Range}
			& \textbf{M1(V)}
			& \textbf{M2(A)}
			& \textbf{Avg} \\
			\midrule

			$\delta_{--}$
			& $[-0.3,-0.1]$
			& \textbf{\underline{0.7785}}
			& 0.7076
			& \textbf{0.7431} \\

			$\delta_{-}$
			& $[-0.2,0.0]$
			& \textbf{0.7750}
			& 0.7007
			& 0.7379 \\

			$\delta_{+}$
			& $[0.0,0.2]$
			& 0.7604
			& \textbf{0.7125}
			& 0.7365 \\

			$\delta_{++}$
			& $[0.1,0.3]$
			& 0.7472
			& 0.7070
			& 0.7271 \\

			\rowcolor{gray!10}
			$\delta$ \textbf{(Ours)}
			& $[-0.1,0.1]$
			& \textbf{0.7750}
			& \textbf{\underline{0.7146}}
			& \textbf{\underline{0.7448}} \\
			\bottomrule
		\end{tabular}
	}

\end{table}

\subsection{Sensitivity to Margin Range and Constraint Relaxation}
\label{appendix:margin_range}

\begin{figure*}[htbp]
	\centering
	\includegraphics[width=0.85\linewidth]{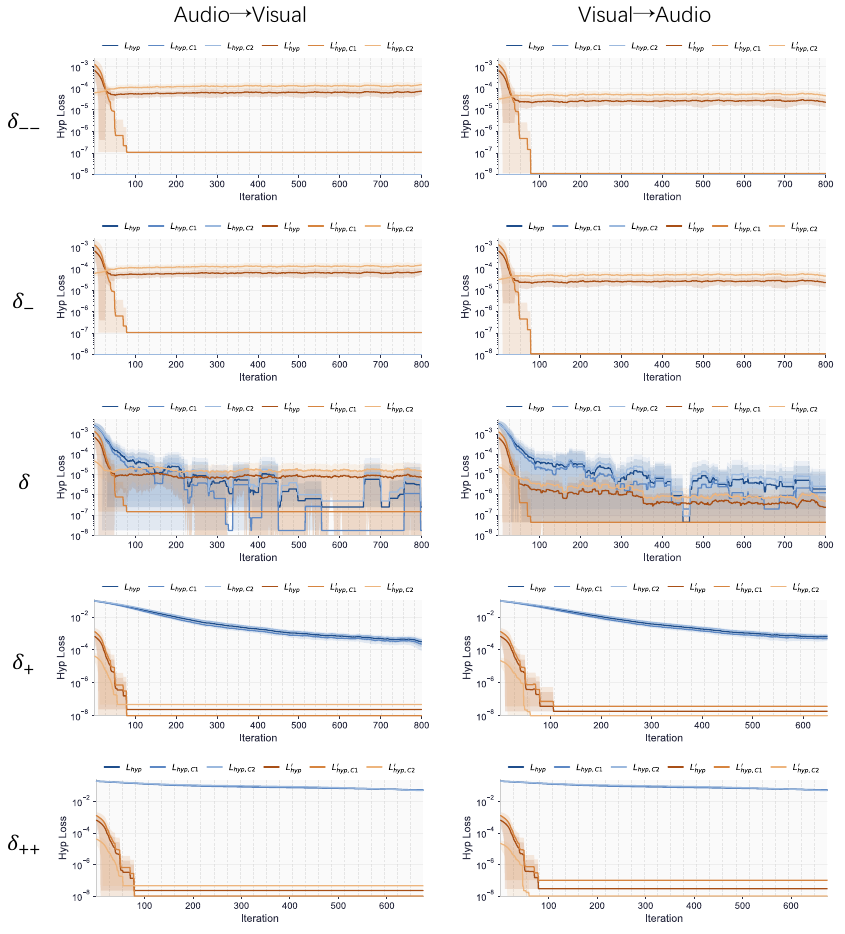}
	\caption{Training dynamics of $\mathcal{L}_\text{hyp}$ and its components under different margin ranges on RAVDESS. Left column: A$\to$V; right column: V$\to$A. Each subplot shows the with-margin losses ($\mathcal{L}_\text{hyp}$, $\mathcal{L}_{\text{hyp},\text{C1}}$, $\mathcal{L}_{\text{hyp},\text{C2}}$) alongside their margin-free counterparts ($\mathcal{L}'_\text{hyp}$, $\mathcal{L}'_{\text{hyp},\text{C1}}$, $\mathcal{L}'_{\text{hyp},\text{C2}}$).}
	\label{fig:margin_range_dynamics}
\end{figure*}

This section provides the empirical basis for the signed margin
design discussed in Sections~\ref{sec:formulation}
and~\ref{sec:criteria_status} of the main text.
Section~\ref{sec:criteria_status} analyzes constraint relaxation
under the default margin range
$[\delta_{\min},\delta_{\max}]=[-0.1,0.1]$.
Here we vary this range to examine its effect on performance
and constraint behavior. Five configurations are compared on
RAVDESS: two non-positive ranges ($\delta_{--}$, $\delta_{-}$),
two non-negative ranges ($\delta_{+}$, $\delta_{++}$),
and the symmetric range $\delta$ (ours), with all other
hyperparameters identical.

\textbf{Performance (Table~\ref{tab:margin_range}).}
The symmetric range $\delta$ achieves the highest Avg (0.7448).
Relative to this setting, $\delta_{--}$ increases M1 accuracy
from 0.7750 to 0.7785 but decreases M2 accuracy from 0.7146
to 0.7076. The non-negative range $\delta_{+}$ yields lower
accuracy in both M1 (0.7604) and M2 (0.7125).
The more positive range $\delta_{++}$ obtains the lowest Avg
(0.7271), with M1 accuracy matching the w/o KD baseline (0.7472).
All four one-sided ranges underperform the symmetric setting
in average accuracy.

\textbf{Training dynamics (Figure~\ref{fig:margin_range_dynamics}).}
We compare the margin-adjusted losses with their margin-free diagnostics, evaluated on the same learned components with $\delta=0$ in the hinge expressions. In the tested non-positive ranges, $\mathcal{L}_{\text{hyp}}$ remains zero during training, leaving $\mathcal{R}_{\text{ratio}}$ and $\mathcal{R}_{\text{aux}}$ to provide the active optimization signals. Positive margins require larger cross-entropy gaps for zero penalty, and the corresponding hypothesis losses remain active. The curves become flatter under the larger positive-margin setting.

Across all five settings, the margin-free diagnostic $\mathcal{L}'_{\text{hyp}}$ reaches comparably low values, indicating small average violations of the zero-margin criteria. These values alone do not account for the differences in downstream accuracy. The same predicted margin also determines the Stage~2 fusion weight through Eq.~\ref{eq:adaptive_weight}, so changing its allowed range affects both the decomposition objective and the construction of the distillation target.

The symmetric default interval allows the Modulator to tighten or relax the required gaps for individual samples. The non-positive and non-negative intervals restrict this flexibility to one side of zero. The higher average accuracy of the symmetric setting supports this choice for the complete pipeline, in which the margin jointly controls criterion penalties and fusion weights.
Frozen final-checkpoint diagnostics under the default range are
reported separately in Section~\ref{appendix:cmp}.

\begin{figure*}[htbp]
	\centering
	\includegraphics[width=\linewidth]{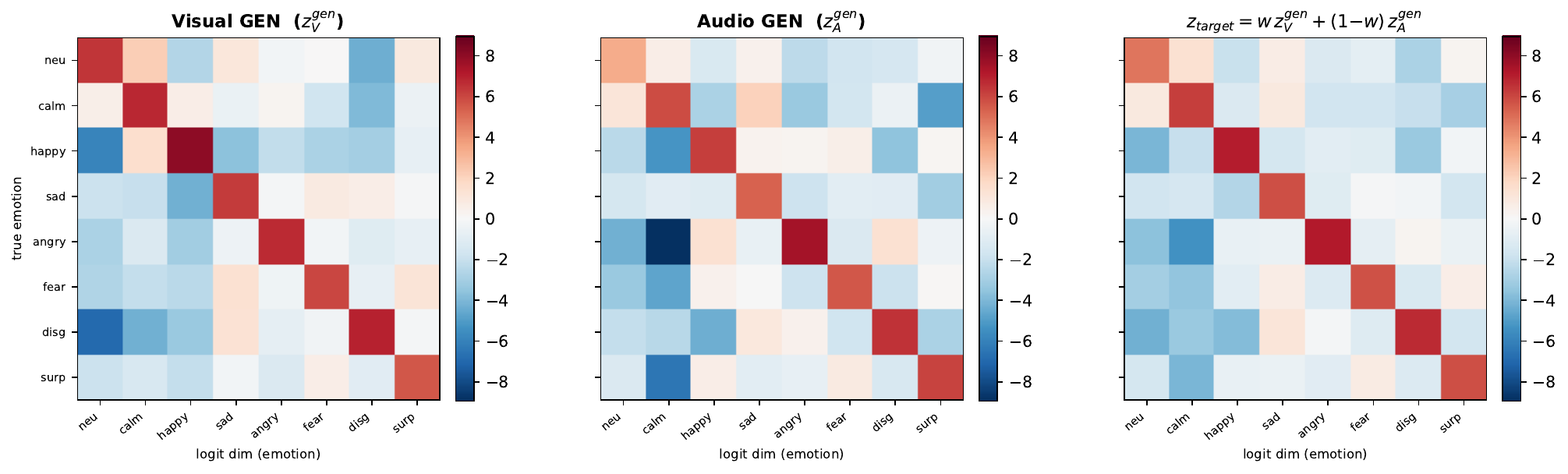}
	\caption{Per-class heatmaps of the modality-general components and the fused distillation target on RAVDESS (V$\to$A), with rows the true emotion and columns the logit dimension. Left and middle: the visual teacher and audio reference-student GEN logits $\mathbf{z}^{gen}_V$ and $\mathbf{z}^{gen}_A$ each keep a strong class-discriminative diagonal but differ in their off-diagonal patterns. Right: the adaptively fused target $\mathbf{z}_{target} = w\,\mathbf{z}^{gen}_V + (1-w)\,\mathbf{z}^{gen}_A$ retains the diagonal while combining both modalities' off-diagonal structure.}
	\label{fig:gen_target}
\end{figure*}

\subsection{Visualization of Disentangled Components}
\label{appendix:disentangle_vis}

\begin{figure}[htbp]
	\centering
	\includegraphics[width=1.0\linewidth]{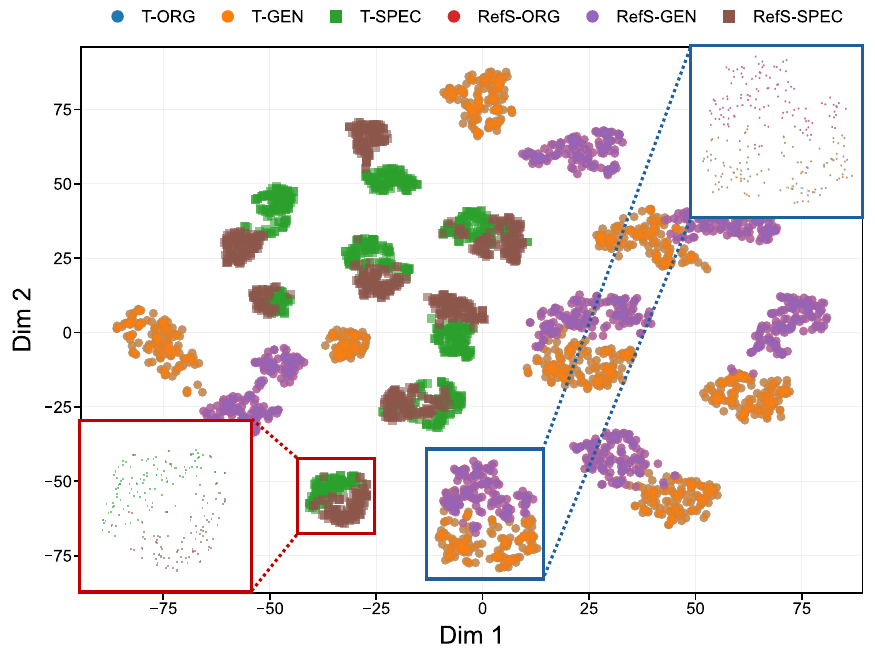}
	\caption{t-SNE visualization of disentangled logit components. T and RefS denote the teacher and reference student; ORG, GEN, and SPEC denote the original logits, modality-general components, and modality-specific components, respectively. All six groups are projected jointly. Red inset: T-SPEC and RefS-SPEC form separated compact clusters. Blue inset: ORG and GEN of the same modality are nearly co-located.}
	\label{fig:disentangle_tsne}
\end{figure}

Figure~\ref{fig:pca_decomposition} provides a PCA view of the residual decomposition. Figure~\ref{fig:disentangle_tsne} complements this view by jointly embedding the six groups of logits on RAVDESS. In the t-SNE projection, T-SPEC and RefS-SPEC form distinct compact clusters with visible class-level organization, while ORG and GEN largely overlap within each modality. These projected patterns complement the per-class component analysis in Section~\ref{sec:semantic_validity} and the quantitative comparison of sharing strategies in Table~\ref{tab:module}.

\begin{figure*}[t]
\begin{minipage}[t]{\columnwidth}
    \vspace{0pt}
	\centering
	\includegraphics[width=1\linewidth]{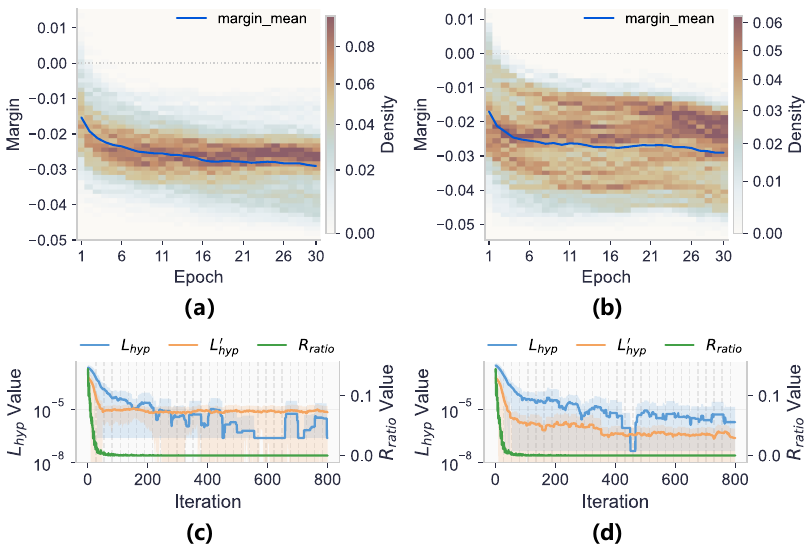}
	\caption{Stage-1 training dynamics from one representative RAVDESS run. (a,b) Per-epoch distributions of the sample-wise predicted margin $\delta$ for A$\to$V and V$\to$A; blue curves denote epoch-wise means. (c,d) Smoothed, log-scaled loss curves, where $\mathcal{L}'_\text{hyp}$ is evaluated with $\delta=0$.}
	\label{fig:training_dynamics}
\end{minipage}\hfill
\begin{minipage}[t]{\columnwidth}
    \vspace{0pt}
	\centering
	\includegraphics[width=1\linewidth]{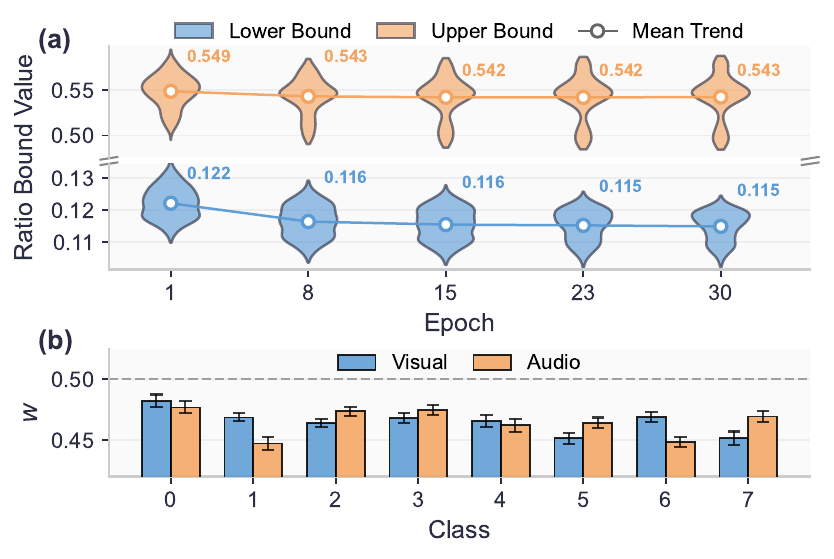}
	\caption{Modulator predictions from one representative RAVDESS run. (a) Sample-wise ratio-bound distributions at selected epochs for A$\to$V; markers and lines denote means. (b) Class-wise mean teacher weights $w$ for both directions; error bars denote standard deviations across training samples, and Visual/Audio indicate the target modality.}
	\label{fig:ratio_weight}
\end{minipage}
\end{figure*}

Figure~\ref{fig:gen_target} shows the class-wise GEN components and their adaptively fused target for V$\to$A. The visual and audio GEN components exhibit prominent diagonals but differ in their off-diagonal patterns. The fused target retains a prominent diagonal while combining these differences. Its closer resemblance to the audio GEN pattern is consistent with the greater average weight assigned to the target modality (Figure~\ref{fig:ratio_weight}(b)).

\subsection{Per-Dataset Efficiency and Accuracy Comparison}
\label{appendix:efficiency}

Table~\ref{tab:efficiency} provides the dataset-level breakdown
underlying Table~\ref{tab:cost_e2e}. Each stage entry is averaged
over both transfer directions. Table~\ref{tab:cost_e2e} first takes
the unweighted mean of each stage across the five datasets, then
sums Params, FLOPs, and Time and takes the maximum stage-averaged
Mem for Total. Table~\ref{tab:efficiency} applies the same total
rules within each dataset; averaging its Total Mem entries need
not reproduce the maximum of the cross-dataset stage means.
For MGDFR, S1, S2, and S3 denote unimodal training, feature-saliency
computation, and student distillation, respectively.
For TALD, S1 and S2 denote logit disentanglement and student distillation.
All recorded stages are included in Total. Small differences
between reported additive totals and sums of displayed entries
can arise from rounding.

TALD reuses its two pretrained unimodal models across opposite
transfer directions (Section~\ref{appendix:training_cfg}). For
additive S0 costs, each direction is allocated half the combined
pretraining cost of the two models, with each model trained once.
Thus, S0 accounts for both models under the bidirectional
evaluation protocol; peak memory follows the stage-wise
aggregation defined above.

The per-dataset records reveal substantial variation in cost.
TALD's disentanglement stage uses 0.0153--0.0180\,M trainable parameters
on four datasets, compared with 5.6962\,M on VGGSound-50k.
FD-CMKD uses 317.5461\,M trainable parameters in S1 on AV-MNIST,
which dominates its cross-dataset mean parameter count.
TALD requires less total computation and time than MGDFR on AV-MNIST,
RAVDESS, VGGSound-50k, and Fakeddit; on CrisisMMD-V2, it reduces
computation but takes longer.
These differences show why the aggregate comparison should be read
alongside the individual dataset results.

The corresponding accuracy gains are listed in
Table~\ref{tab:accuracy_gains} and discussed alongside
positive-transfer frequencies in
Section~\ref{appendix:positive_transfer_frequency}.

\begin{table}[t]
	\caption{Accuracy gains (percentage points) over w/o KD, averaged
		over both transfer directions and five runs.
		Avg denotes the unweighted mean across D1--D5 and corresponds to
		$\Delta$Acc in Table~\ref{tab:cm_performance}.
		Blue and red denote positive and negative gains, respectively;
		shading intensity increases logarithmically with absolute gain.}
	\label{tab:accuracy_gains}
	\centering

	\begingroup
	\fontsize{8.5}{10}\selectfont
	\rmfamily
	\renewcommand{\arraystretch}{1.08}
	\setlength{\tabcolsep}{1.5pt}

	\definecolor{GainPositive}{RGB}{44,127,184}
	\definecolor{GainNegative}{RGB}{214,96,77}

	\edef\GainColWidth{%
		\the\dimexpr(\columnwidth-40pt-14\tabcolsep)/6\relax}

	\ExplSyntaxOn
	\cs_set_protected:Npn \GainCell #1
	{
		\fp_compare:nNnTF {#1} = {0}
		{ \cellcolor{white} }
		{
			\fp_set:Nn \l_tmpa_fp
			{
				round(
				min(60, 8 + 52 * ln(1 + abs(#1)) / ln(31.86)),
				1
				)
			}
			\fp_compare:nNnTF {#1} > {0}
			{
				\exp_args:Nx \cellcolor
				{ GainPositive!\fp_to_decimal:N \l_tmpa_fp }
			}
			{
				\exp_args:Nx \cellcolor
				{ GainNegative!\fp_to_decimal:N \l_tmpa_fp }
			}
		}
		\ensuremath{#1}
	}
	\ExplSyntaxOff

	\begin{tabular}{
			>{\raggedright\arraybackslash}m{40pt}
			*{6}{>{\centering\arraybackslash}m{\GainColWidth}}
		}
		\toprule
		\textbf{Method}
		& \textbf{D1} & \textbf{D2} & \textbf{D3}
		& \textbf{D4} & \textbf{D5} & \textbf{Avg} \\
		\midrule

		\multicolumn{7}{l}{\textit{Single-Modal}} \\

		KD
		& \GainCell{+0.06} & \GainCell{-0.93} & \GainCell{-0.78}
		& \GainCell{-0.25} & \GainCell{-0.90} & \GainCell{-0.56} \\

		MLLD
		& \GainCell{+0.07} & \GainCell{-0.80} & \GainCell{+0.22}
		& \GainCell{-1.30} & \GainCell{-12.94} & \GainCell{-2.95} \\

		FitNets
		& \GainCell{-1.00} & \GainCell{-5.24} & \GainCell{-2.35}
		& \GainCell{-1.53} & \GainCell{+0.16} & \GainCell{-1.99} \\

		OFA
		& \GainCell{+0.10} & \GainCell{-2.15} & \GainCell{-0.99}
		& \GainCell{-0.08} & \GainCell{+0.66} & \GainCell{-0.49} \\

		RKD
		& \GainCell{-2.53} & \GainCell{+1.15} & \GainCell{-0.48}
		& \GainCell{-0.01} & \GainCell{-17.53} & \GainCell{-3.88} \\

		CRD
		& \GainCell{-0.12} & \GainCell{+2.92} & \GainCell{-2.67}
		& \GainCell{-0.85} & \GainCell{-7.39} & \GainCell{-1.62} \\
		\midrule

		\multicolumn{7}{l}{\textit{Cross-Modal}} \\

		MGDFR
		& \GainCell{+0.12} & \GainCell{+2.61} & \GainCell{+2.22}
		& \GainCell{-0.35} & \GainCell{+0.40} & \GainCell{+1.00} \\

		C$^2$KD
		& \GainCell{+0.02} & \GainCell{+0.44} & \GainCell{+2.06}
		& \GainCell{-0.99} & \GainCell{-30.86} & \GainCell{-5.87} \\

		FD-CMKD
		& \GainCell{-0.18} & \GainCell{+1.67} & \GainCell{+1.44}
		& \GainCell{+0.02} & \GainCell{-0.14} & \GainCell{+0.56} \\

		\textbf{TALD}
		& \GainCell{+0.09} & \GainCell{+3.65} & \GainCell{+2.63}
		& \GainCell{+0.33} & \GainCell{+0.21} & \GainCell{+1.38} \\
		\bottomrule
	\end{tabular}

	\vspace{2pt}
	\parbox{\columnwidth}{%
		\fontsize{7.5}{8.5}\selectfont\raggedright
		D1: AV-MNIST; D2: RAVDESS; D3: VGGSound-50k;\\
		D4: CrisisMMD-V2; D5: Fakeddit.}
	\endgroup
\end{table}

\newpage
\subsection{Complete Training Dynamics}
\label{appendix:full_dynamics}

Figures~\ref{fig:training_dynamics} and~\ref{fig:ratio_weight} provide the complete Stage-1 dynamics from one RAVDESS run and the sample-level evidence summarized in Section~\ref{sec:criteria_status}.
These epoch-wise training observations are distinct from the
frozen-checkpoint test-set diagnostics in Section~\ref{appendix:cmp}.

\paragraph{Margin dynamics and criterion violations.}
The predicted margins retain sample-level variation and evolve differently across directions (Figure~\ref{fig:training_dynamics}(a,b)). The diagnostic $\mathcal{L}'_{\text{hyp}}$ is evaluated on the same learned components as $\mathcal{L}_{\text{hyp}}$ with $\delta=0$ in the hinge expressions. In V$\to$A, $\mathcal{L}_{\text{hyp}}$ remains above $\mathcal{L}'_{\text{hyp}}$ while both decrease. In A$\to$V, the mean margin becomes negative and $\mathcal{L}'_{\text{hyp}}$ plateaus as $\mathcal{L}_{\text{hyp}}$ continues to decline. The average zero-margin violation therefore remains stable during this later decrease in the margin-adjusted loss.

\paragraph{Ratio bounds.}
$\mathcal{R}_{\text{ratio}}$ decreases in both directions (Figure~\ref{fig:training_dynamics}(c,d)). For A$\to$V, Figure~\ref{fig:ratio_weight}(a) shows that both predicted bounds retain within-epoch variation and remain separated throughout the observed run.

\paragraph{Adaptive weights.}
Figure~\ref{fig:ratio_weight}(b) reports class-wise mean teacher weights and standard deviations across training samples. The weights vary across classes and transfer directions, while all class-wise means remain below 0.5, indicating a higher average fusion weight for the reference student's modality-general component.

\onecolumn
\begin{landscape}
\begin{table}[p]
	\centering
	\caption{Complete stage-wise and end-to-end cost records on five
		benchmarks, with stage entries averaged over both transfer directions.
		Params, FLOPs, Mem, and Time are reported in M, T, MiB, and seconds,
		respectively. Stage labels follow Table~\ref{tab:cost_e2e}.
		Within each dataset, Total sums Params, FLOPs, and Time and takes
		the maximum of the displayed stage Mem values.}
	\label{tab:efficiency}

	\begingroup
	\fontsize{8}{9.5}\selectfont
	\rmfamily
	\setlength{\tabcolsep}{2pt}
	\renewcommand{\arraystretch}{1.05}
	\edef\CompactHeaderShift{%
		\the\dimexpr0pt-(0.4ex+\cmidrulewidth)/2\relax}

	\resizebox{0.98\linewidth}{!}{%
		\begin{tabular}{@{}cl*{5}{rrrr}@{}}
			\toprule
			\multirow[c]{2}{*}[\CompactHeaderShift]{\textbf{Method}}
			& \multirow[c]{2}{*}[\CompactHeaderShift]{\textbf{Stage}}
			& \multicolumn{4}{c}{\textbf{AV-MNIST}}
			& \multicolumn{4}{c}{\textbf{RAVDESS}}
			& \multicolumn{4}{c}{\textbf{VGGSound-50k}}
			& \multicolumn{4}{c}{\textbf{CrisisMMD-V2}}
			& \multicolumn{4}{c}{\textbf{Fakeddit}} \\
			\noalign{\vskip\dimexpr0.2ex-\aboverulesep\relax}
			\cmidrule(lr){3-6}
			\cmidrule(lr){7-10}
			\cmidrule(lr){11-14}
			\cmidrule(lr){15-18}
			\cmidrule(lr){19-22}
			\noalign{\vskip\dimexpr0.2ex-\belowrulesep\relax}
			& & \textbf{Params} & \textbf{FLOPs} & \textbf{Mem} & \textbf{Time}
			& \textbf{Params} & \textbf{FLOPs} & \textbf{Mem} & \textbf{Time}
			& \textbf{Params} & \textbf{FLOPs} & \textbf{Mem} & \textbf{Time}
			& \textbf{Params} & \textbf{FLOPs} & \textbf{Mem} & \textbf{Time}
			& \textbf{Params} & \textbf{FLOPs} & \textbf{Mem} & \textbf{Time} \\
			\midrule

			& S0
			& 0.2026 & 1500.89 & 3331.02 & 645.71
			& 3.3301 & 134.39 & 2594.12 & 373.54
			& 2.4632 & 10507.31 & 1578.35 & 1353.95
			& 1.9717 & 8.66 & 75.38 & 440.56
			& 1.9707 & 26.95 & 75.36 & 451.50 \\
			& S1
			& 0.2026 & 2007.40 & 3333.50 & 684.78
			& 3.3301 & 189.35 & 2596.61 & 632.18
			& 2.4632 & 15700.40 & 1583.22 & 2096.39
			& 1.9717 & 12.46 & 84.19 & 457.07
			& 1.9707 & 38.76 & 83.05 & 482.74 \\
			\rowcolor{gray!12}
			\cellcolor{white}\multirow[c]{-3}{*}{KD}
			& Total
			& 0.4053 & 3508.28 & 3333.50 & 1330.49
			& 6.6603 & 323.74 & 2596.61 & 1005.73
			& 4.9263 & 26207.71 & 1583.22 & 3450.34
			& 3.9434 & 21.12 & 84.19 & 897.63
			& 3.9414 & 65.71 & 83.05 & 934.23 \\
			\midrule

			& S0
			& 0.2026 & 1500.89 & 3331.02 & 645.71
			& 3.3301 & 134.39 & 2594.12 & 373.54
			& 2.4632 & 10507.31 & 1578.35 & 1353.95
			& 1.9717 & 8.66 & 75.38 & 440.56
			& 1.9707 & 26.95 & 75.36 & 451.50 \\
			& S1
			& 0.2026 & 2008.35 & 3333.50 & 713.96
			& 3.3301 & 189.35 & 2596.61 & 623.16
			& 2.4632 & 15711.32 & 1583.22 & 2005.57
			& 1.9717 & 12.64 & 84.62 & 465.81
			& 1.9707 & 39.19 & 84.51 & 107.18 \\
			\rowcolor{gray!12}
			\cellcolor{white}\multirow[c]{-3}{*}{MLLD}
			& Total
			& 0.4053 & 3509.23 & 3333.50 & 1359.67
			& 6.6603 & 323.74 & 2596.61 & 996.71
			& 4.9263 & 26218.63 & 1583.22 & 3359.52
			& 3.9434 & 21.30 & 84.62 & 906.37
			& 3.9414 & 66.14 & 84.51 & 558.68 \\
			\midrule

			& S0
			& 0.2026 & 1500.89 & 3331.02 & 645.71
			& 3.3301 & 134.39 & 2594.12 & 373.54
			& 2.4632 & 10507.31 & 1578.35 & 1353.95
			& 1.9717 & 8.66 & 75.38 & 440.56
			& 1.9707 & 26.95 & 75.36 & 451.50 \\
			& S1
			& 0.0128 & 2077.72 & 6555.53 & 750.78
			& 3.2670 & 189.35 & 3943.01 & 574.17
			& 0.1131 & 15699.99 & 2348.48 & 2115.55
			& 1.9676 & 12.49 & 89.61 & 660.34
			& 1.9676 & 38.85 & 89.60 & 490.02 \\
			& S2
			& 0.2026 & 2007.40 & 3827.43 & 684.78
			& 3.3301 & 189.35 & 3378.89 & 650.95
			& 2.4632 & 15700.40 & 1602.67 & 2263.00
			& 1.9717 & 12.46 & 115.39 & 450.67
			& 1.9707 & 38.76 & 119.14 & 483.63 \\
			\rowcolor{gray!12}
			\cellcolor{white}\multirow[c]{-4}{*}{FitNets}
			& Total
			& 0.4181 & 5586.00 & 6555.53 & 2081.28
			& 9.9273 & 513.08 & 3943.01 & 1598.66
			& 5.0395 & 41907.70 & 2348.48 & 5732.51
			& 5.9111 & 33.61 & 115.39 & 1551.56
			& 5.9090 & 104.57 & 119.14 & 1425.15 \\
			\midrule

			& S0
			& 0.2026 & 1500.89 & 3331.02 & 645.71
			& 3.3301 & 134.39 & 2594.12 & 373.54
			& 2.4632 & 10507.31 & 1578.35 & 1353.95
			& 1.9717 & 8.66 & 75.38 & 440.56
			& 1.9707 & 26.95 & 75.36 & 451.50 \\
			& S1
			& 0.2933 & 4043.67 & 7459.85 & 927.07
			& 3.4990 & 189.44 & 3944.32 & 543.65
			& 2.7732 & 15705.87 & 2353.51 & 2257.13
			& 1.9966 & 12.61 & 89.15 & 456.91
			& 1.9893 & 39.11 & 89.27 & 69.00 \\
			\rowcolor{gray!12}
			\cellcolor{white}\multirow[c]{-3}{*}{OFA}
			& Total
			& 0.4959 & 5544.56 & 7459.85 & 1572.78
			& 6.8292 & 323.83 & 3944.32 & 917.19
			& 5.2364 & 26213.19 & 2353.51 & 3611.08
			& 3.9683 & 21.27 & 89.15 & 897.47
			& 3.9600 & 66.06 & 89.27 & 520.49 \\
			\midrule

			& S0
			& 0.2026 & 1500.89 & 3331.02 & 645.71
			& 3.3301 & 134.39 & 2594.12 & 373.54
			& 2.4632 & 10507.31 & 1578.35 & 1353.95
			& 1.9717 & 8.66 & 75.38 & 440.56
			& 1.9707 & 26.95 & 75.36 & 451.50 \\
			& S1
			& 3.4249 & 1601.79 & 6835.06 & 390.99
			& 3.3714 & 94.73 & 3946.49 & 259.35
			& 2.5044 & 8247.25 & 3774.68 & 1314.42
			& 2.1030 & 135.58 & 2652.19 & 235.90
			& 2.1020 & 425.25 & 2652.16 & 280.88 \\
			\rowcolor{gray!12}
			\cellcolor{white}\multirow[c]{-3}{*}{RKD}
			& Total
			& 3.6275 & 3102.68 & 6835.06 & 1036.70
			& 6.7015 & 229.12 & 3946.49 & 632.90
			& 4.9676 & 18754.57 & 3774.68 & 2668.37
			& 4.0748 & 144.25 & 2652.19 & 676.46
			& 4.0727 & 452.20 & 2652.16 & 732.38 \\
			\midrule

			& S0
			& 0.2026 & 1500.89 & 3331.02 & 645.71
			& 3.3301 & 134.39 & 2594.12 & 373.54
			& 2.4632 & 10507.31 & 1578.35 & 1353.95
			& 1.9717 & 8.66 & 75.38 & 440.56
			& 1.9707 & 26.95 & 75.36 & 451.50 \\
			& S1
			& 3.4249 & 2093.88 & 6929.37 & 1046.70
			& 3.3714 & 189.74 & 3948.61 & 610.58
			& 2.5044 & 15713.71 & 3391.96 & 2537.64
			& 2.1030 & 17.29 & 2258.20 & 487.11
			& 2.1020 & 53.78 & 2279.53 & 1576.00 \\
			\rowcolor{gray!12}
			\cellcolor{white}\multirow[c]{-3}{*}{CRD}
			& Total
			& 3.6275 & 3594.76 & 6929.37 & 1692.41
			& 6.7015 & 324.13 & 3948.61 & 984.13
			& 4.9676 & 26221.02 & 3391.96 & 3891.60
			& 4.0748 & 25.95 & 2258.20 & 927.67
			& 4.0727 & 80.73 & 2279.53 & 2027.50 \\
			\midrule

			& S1
			& 0.4053 & 3001.74 & 3348.90 & 794.34
			& 6.6603 & 268.59 & 2591.45 & 561.07
			& 4.9263 & 21014.34 & 1583.96 & 2049.82
			& 3.9434 & 17.25 & 78.83 & 461.55
			& 3.9414 & 53.67 & 78.79 & 498.02 \\
			& S2
			& 0.0000 & 6480.52 & 1616.83 & 652.47
			& 0.0000 & 879.95 & 1473.83 & 300.56
			& 0.0000 & 83141.13 & 1063.77 & 2867.53
			& 0.0000 & 194.48 & 79.92 & 39.86
			& 0.0000 & 605.06 & 80.79 & 211.53 \\
			& S3
			& 0.2026 & 2007.55 & 5392.03 & 697.45
			& 3.3301 & 189.35 & 3997.03 & 622.96
			& 2.4632 & 15700.40 & 2387.18 & 2094.14
			& 1.9717 & 12.46 & 126.11 & 446.82
			& 1.9707 & 38.77 & 128.16 & 508.72 \\
			\rowcolor{gray!12}
			\cellcolor{white}\multirow[c]{-4}{*}{MGDFR}
			& Total
			& 0.6079 & 11489.81 & 5392.03 & 2144.25
			& 9.9904 & 1337.90 & 3997.03 & 1484.58
			& 7.3895 & 119855.87 & 2387.18 & 7011.49
			& 5.9152 & 224.19 & 126.11 & 948.23
			& 5.9121 & 697.49 & 128.16 & 1218.26 \\
			\midrule

			& S0
			& 0.2026 & 1500.89 & 3331.02 & 645.71
			& 3.3301 & 134.39 & 2594.12 & 373.54
			& 2.4632 & 10507.31 & 1578.35 & 1353.95
			& 1.9717 & 8.66 & 75.38 & 440.56
			& 1.9707 & 26.95 & 75.36 & 451.50 \\
			& S1
			& 0.4314 & 3002.72 & 3398.56 & 844.28
			& 6.8721 & 268.70 & 2584.00 & 597.42
			& 5.2236 & 21019.77 & 1599.53 & 2022.51
			& 4.4790 & 20.36 & 93.05 & 468.96
			& 4.4749 & 63.30 & 95.64 & 539.51 \\
			\rowcolor{gray!12}
			\cellcolor{white}\multirow[c]{-3}{*}{C$^2$KD}
			& Total
			& 0.6340 & 4503.61 & 3398.56 & 1489.99
			& 10.2023 & 403.09 & 2594.12 & 970.96
			& 7.6868 & 31527.09 & 1599.53 & 3376.46
			& 6.4507 & 29.02 & 93.05 & 909.52
			& 6.4456 & 90.25 & 95.64 & 991.01 \\
			\midrule

			& S0
			& 0.2026 & 1500.89 & 3331.02 & 645.71
			& 3.3301 & 134.39 & 2594.12 & 373.54
			& 2.4632 & 10507.31 & 1578.35 & 1353.95
			& 1.9717 & 8.66 & 75.38 & 440.56
			& 1.9707 & 26.95 & 75.36 & 451.50 \\
			& S1
			& 317.5461 & 7385.94 & 8213.04 & 968.38
			& 3.3353 & 189.36 & 2596.74 & 560.89
			& 2.5540 & 15702.65 & 1581.16 & 1811.63
			& 1.9881 & 12.74 & 92.71 & 496.56
			& 1.9830 & 39.52 & 92.56 & 556.81 \\
			\rowcolor{gray!12}
			\cellcolor{white}\multirow[c]{-3}{*}{FD-CMKD}
			& Total
			& 317.7487 & 8886.83 & 8213.04 & 1614.09
			& 6.6654 & 323.74 & 2596.74 & 934.43
			& 5.0171 & 26209.96 & 1581.16 & 3165.59
			& 3.9599 & 21.40 & 92.71 & 937.12
			& 3.9537 & 66.47 & 92.56 & 1008.30 \\
			\midrule

			& S0
			& 0.2026 & 1500.89 & 3331.02 & 645.71
			& 3.3301 & 134.39 & 2594.12 & 373.54
			& 2.4632 & 10507.31 & 1578.35 & 1353.95
			& 1.9717 & 8.66 & 75.38 & 440.56
			& 1.9707 & 26.95 & 75.36 & 451.50 \\
			& S1
			& 0.0180 & 304.33 & 1642.40 & 188.71
			& 0.0165 & 32.98 & 1999.16 & 194.71
			& 5.6962 & 3116.73 & 1710.38 & 638.93
			& 0.0165 & 2.37 & 68.30 & 154.38
			& 0.0153 & 7.39 & 66.11 & 180.13 \\
			& S2
			& 0.2026 & 2514.09 & 3361.08 & 731.26
			& 3.3301 & 244.32 & 2896.27 & 731.92
			& 2.4632 & 20893.90 & 2138.19 & 2104.80
			& 1.9717 & 16.30 & 97.44 & 492.61
			& 1.9707 & 50.71 & 97.41 & 519.10 \\
			\rowcolor{gray!12}
			\cellcolor{white}\multirow[c]{-4}{*}{TALD}
			& Total
			& 0.4233 & 4319.31 & 3361.08 & 1565.67
			& 6.6767 & 411.68 & 2896.27 & 1300.17
			& 10.6226 & 34517.94 & 2138.19 & 4097.68
			& 3.9599 & 27.34 & 97.44 & 1087.55
			& 3.9567 & 85.04 & 97.41 & 1150.73 \\
			\bottomrule
		\end{tabular}%
	}
	\endgroup
\end{table}
\end{landscape}

\definecolor{TransferPositive}{RGB}{44,127,184}
\ExplSyntaxOn
\cs_set_protected:Npn \PositiveRateCell #1#2
{
	\fp_set:Nn \l_tmpa_fp { round(0.6 * (#1), 1) }
	\exp_args:Nx \cellcolor
	{ TransferPositive!\fp_to_decimal:N \l_tmpa_fp }
	#2
}
\ExplSyntaxOff
\twocolumn[{
	\begin{minipage}{\textwidth}
		\captionsetup{type=table}
		\caption{Positive-transfer counts by transfer direction. Each entry reports the number of runs with strictly positive accuracy gains over the corresponding w/o KD baseline out of five ($k^{+}/5$). PTF matches the overall frequency reported in Table~\ref{tab:cm_performance}. Arrows indicate teacher-to-student transfer; I, V, A, and T denote image, visual, audio, and text, respectively. Blue shading encodes the corresponding percentages on a shared 0--100\% scale. The highest PTF is bold.}
		\label{tab:positive_transfer_frequency}
		\centering
		\begingroup
		\fontsize{9}{11}\selectfont
		\rmfamily
		\renewcommand{\arraystretch}{1.08}
		\setlength{\tabcolsep}{8pt}
		\edef\CompactHeaderShift{%
			\the\dimexpr0pt-(0.4ex+\cmidrulewidth)/2\relax}

		\begin{tabular}{l*{11}{c}}
			\toprule
			\multirow[c]{2}{*}[\CompactHeaderShift]{\textbf{Method}}
			& \multicolumn{2}{c}{\textbf{AV-MNIST}}
			& \multicolumn{2}{c}{\textbf{RAVDESS}}
			& \multicolumn{2}{c}{\textbf{VGGSound-50k}}
			& \multicolumn{2}{c}{\textbf{CrisisMMD-V2}}
			& \multicolumn{2}{c}{\textbf{Fakeddit}}
			& \multirow[c]{2}{*}[\CompactHeaderShift]{\textbf{PTF}} \\
			\noalign{\vskip\dimexpr0.2ex-\aboverulesep\relax}
			\cmidrule(lr){2-3}\cmidrule(lr){4-5}\cmidrule(lr){6-7}
			\cmidrule(lr){8-9}\cmidrule(lr){10-11}
			\noalign{\vskip\dimexpr0.2ex-\belowrulesep\relax}
			& $\mathrm{A}\!\to\!\mathrm{I}$ & $\mathrm{I}\!\to\!\mathrm{A}$
			& $\mathrm{A}\!\to\!\mathrm{V}$ & $\mathrm{V}\!\to\!\mathrm{A}$
			& $\mathrm{A}\!\to\!\mathrm{V}$ & $\mathrm{V}\!\to\!\mathrm{A}$
			& $\mathrm{T}\!\to\!\mathrm{I}$ & $\mathrm{I}\!\to\!\mathrm{T}$
			& $\mathrm{T}\!\to\!\mathrm{I}$ & $\mathrm{I}\!\to\!\mathrm{T}$ & \\
			\midrule
			KD & \PositiveRateCell{60}{3/5} & \PositiveRateCell{60}{3/5} & \PositiveRateCell{0}{0/5} & \PositiveRateCell{80}{4/5} & \PositiveRateCell{0}{0/5} & \PositiveRateCell{100}{5/5} & \PositiveRateCell{40}{2/5} & \PositiveRateCell{20}{1/5} & \PositiveRateCell{60}{3/5} & \PositiveRateCell{0}{0/5} & \PositiveRateCell{42}{42\%} \\
			MLLD & \PositiveRateCell{40}{2/5} & \PositiveRateCell{80}{4/5} & \PositiveRateCell{20}{1/5} & \PositiveRateCell{60}{3/5} & \PositiveRateCell{20}{1/5} & \PositiveRateCell{100}{5/5} & \PositiveRateCell{80}{4/5} & \PositiveRateCell{0}{0/5} & \PositiveRateCell{60}{3/5} & \PositiveRateCell{0}{0/5} & \PositiveRateCell{46}{46\%} \\
			\midrule
			FitNets & \PositiveRateCell{0}{0/5} & \PositiveRateCell{0}{0/5} & \PositiveRateCell{0}{0/5} & \PositiveRateCell{0}{0/5} & \PositiveRateCell{0}{0/5} & \PositiveRateCell{0}{0/5} & \PositiveRateCell{0}{0/5} & \PositiveRateCell{0}{0/5} & \PositiveRateCell{0}{0/5} & \PositiveRateCell{100}{5/5} & \PositiveRateCell{10}{10\%} \\
			OFA & \PositiveRateCell{80}{4/5} & \PositiveRateCell{100}{5/5} & \PositiveRateCell{20}{1/5} & \PositiveRateCell{20}{1/5} & \PositiveRateCell{0}{0/5} & \PositiveRateCell{60}{3/5} & \PositiveRateCell{0}{0/5} & \PositiveRateCell{80}{4/5} & \PositiveRateCell{40}{2/5} & \PositiveRateCell{100}{5/5} & \PositiveRateCell{50}{50\%} \\
			\midrule
			RKD & \PositiveRateCell{100}{5/5} & \PositiveRateCell{0}{0/5} & \PositiveRateCell{40}{2/5} & \PositiveRateCell{60}{3/5} & \PositiveRateCell{60}{3/5} & \PositiveRateCell{20}{1/5} & \PositiveRateCell{40}{2/5} & \PositiveRateCell{60}{3/5} & \PositiveRateCell{40}{2/5} & \PositiveRateCell{0}{0/5} & \PositiveRateCell{42}{42\%} \\
			CRD & \PositiveRateCell{0}{0/5} & \PositiveRateCell{80}{4/5} & \PositiveRateCell{80}{4/5} & \PositiveRateCell{80}{4/5} & \PositiveRateCell{0}{0/5} & \PositiveRateCell{60}{3/5} & \PositiveRateCell{40}{2/5} & \PositiveRateCell{0}{0/5} & \PositiveRateCell{40}{2/5} & \PositiveRateCell{0}{0/5} & \PositiveRateCell{38}{38\%} \\
			\midrule
			MGDFR & \PositiveRateCell{60}{3/5} & \PositiveRateCell{100}{5/5} & \PositiveRateCell{60}{3/5} & \PositiveRateCell{100}{5/5} & \PositiveRateCell{100}{5/5} & \PositiveRateCell{100}{5/5} & \PositiveRateCell{0}{0/5} & \PositiveRateCell{20}{1/5} & \PositiveRateCell{100}{5/5} & \PositiveRateCell{60}{3/5} & \PositiveRateCell{70}{70\%} \\
			C\textsuperscript{2}KD & \PositiveRateCell{20}{1/5} & \PositiveRateCell{80}{4/5} & \PositiveRateCell{40}{2/5} & \PositiveRateCell{40}{2/5} & \PositiveRateCell{100}{5/5} & \PositiveRateCell{100}{5/5} & \PositiveRateCell{60}{3/5} & \PositiveRateCell{0}{0/5} & \PositiveRateCell{0}{0/5} & \PositiveRateCell{0}{0/5} & \PositiveRateCell{44}{44\%} \\
			FD-CMKD & \PositiveRateCell{0}{0/5} & \PositiveRateCell{100}{5/5} & \PositiveRateCell{60}{3/5} & \PositiveRateCell{60}{3/5} & \PositiveRateCell{100}{5/5} & \PositiveRateCell{0}{0/5} & \PositiveRateCell{60}{3/5} & \PositiveRateCell{20}{1/5} & \PositiveRateCell{40}{2/5} & \PositiveRateCell{20}{1/5} & \PositiveRateCell{46}{46\%} \\
			\textbf{TALD} & \PositiveRateCell{60}{3/5} & \PositiveRateCell{100}{5/5} & \PositiveRateCell{80}{4/5} & \PositiveRateCell{100}{5/5} & \PositiveRateCell{100}{5/5} & \PositiveRateCell{100}{5/5} & \PositiveRateCell{80}{4/5} & \PositiveRateCell{80}{4/5} & \PositiveRateCell{100}{5/5} & \PositiveRateCell{40}{2/5} & \PositiveRateCell{84}{\textbf{84\%}} \\
			\bottomrule
		\end{tabular}
		\endgroup
		\par\vspace{10pt}
	\end{minipage}
}]

\subsection{Accuracy Gains and Positive-Transfer Frequency}
\label{appendix:positive_transfer_frequency}

Tables~\ref{tab:accuracy_gains} and~\ref{tab:positive_transfer_frequency}
provide detailed breakdowns of the $\Delta$Acc and PTF summaries in
Table~\ref{tab:cm_performance}, using the same five runs per setting.
The former averages accuracy gains over both transfer directions
within each dataset; the latter reports positive-gain counts for
each direction and the overall PTF.

\paragraph{Metric.}
For method $m$ and dataset--direction setting $d$, let $k^{+}_{m,d}$
count the runs with a strictly positive test-accuracy gain over the
corresponding w/o KD baseline among the five runs.
Per-direction entries display $k^{+}_{m,d}/5$.
The corresponding percentage and its unweighted average across
directions are
\begin{equation*}
	p^{+}_{m,d}=\frac{100\,k^{+}_{m,d}}{5},
	\qquad
	\overline{p}^{+}_{m}=\frac{1}{10}
	\sum_{d=1}^{10}p^{+}_{m,d}.
\end{equation*}
The PTF columns in Tables~\ref{tab:cm_performance}
and~\ref{tab:positive_transfer_frequency} report
$\overline{p}^{+}_{m}$, equivalently the percentage of positive gains
across all 50 evaluations.
Ties remain in the denominator but are not counted as positive.

\newpage
\paragraph{Joint results.}
TALD's PTF of 84\% corresponds to 42 positive-gain evaluations out of
50, compared with 70\% (35/50) for MGDFR.
Table~\ref{tab:accuracy_gains} further shows positive mean gains for
TALD on all five datasets.
The directional breakdown provides additional context: on Fakeddit,
TALD's mean gain across both directions is +0.21 percentage points,
while positive transfer occurs in 5/5 runs for T$\to$I and 2/5 for
I$\to$T.
Thus, a positive dataset average can coexist with substantial
directional variation in positive-transfer frequency.

\paragraph{Interpretation.}
PTF measures how often positive gains occur, without accounting for
their magnitudes or class-wise effects.
Nonpositive outcomes include both losses and ties.
The aggregate is a descriptive summary across datasets and directions,
rather than a significance test based on 50 independent trials.

\end{document}